\documentclass[journal,compsoc,onecolumn]{IEEEtran}

\usepackage{graphicx}
\usepackage{amsmath}
\usepackage{multirow}

\usepackage[hidelinks]{hyperref}
\hypersetup{breaklinks=true}
\begin{document}

\title{FastForest: Increasing Random Forest Processing Speed While Maintaining Accuracy}


\author{Darren~Yates
        and~Md~Zahidul~Islam
\IEEEcompsocitemizethanks{\IEEEcompsocthanksitem D. Yates and M.Z. Islam are with the School of
Computing and Mathematics, Charles Sturt University, Panorama Ave, Bathurst, NSW 2795, Australia.\protect\\
E-mail: \{dyates, zislam\}@csu.edu.au}
}





\IEEEtitleabstractindextext{
\begin{abstract}
\noindent Random Forest remains one of Data Mining's most enduring ensemble algorithms, achieving well-documented levels of accuracy and processing speed, as well as regularly appearing in new research. However, with data mining now reaching the domain of hardware-constrained devices such as smartphones and Internet of Things (IoT) devices, there is continued need for further research into algorithm efficiency to deliver greater processing speed without sacrificing accuracy. Our proposed FastForest algorithm delivers an average 24\% increase in processing speed compared with Random Forest whilst maintaining (and frequently exceeding) it on classification accuracy over tests involving 45 datasets. FastForest achieves this result through a combination of three optimising components - Subsample Aggregating (`Subbagging'), Logarithmic Split-Point Sampling and Dynamic Restricted Subspacing. Moreover, detailed testing of Subbagging sizes has found an optimal scalar delivering a positive mix of processing performance and accuracy. 
\end{abstract}
}

\maketitle
\IEEEdisplaynontitleabstractindextext
\IEEEpeerreviewmaketitle
\IEEEraisesectionheading{\section{Introduction}\label{sec:introduction}}

 
\IEEEPARstart{E}{nsemble} classifiers have a rich history in data mining and continue to feature in new research covering topics as varied as malware detection \cite{RN519}, gene selection \cite{RN520} and energy consumption of buildings \cite{RN529}. Their ability to identify multiple information patterns within datasets has seen their popularity continue to grow. Well-known ensemble options include Bootstrapped Aggregating (or ‘Bagging’) \cite{RN455}, which samples the dataset with replacement to create multiple dataset derivatives and from which, multiple decision trees can be created. Random Subspacing \cite{RN456}, another popular alternative, reduces the number of attributes available for selection at a particular tree node through random selection to boost diversity. \\
\indent However, it is Random Forest \cite{RN454}, the algorithm that essentially combines Bagging with Random Subspacing, that continues to feature heavily in new research \cite{RN544,RN545,RN546,RN547,RN548}. This is thanks largely to its ability to achieve high levels of accuracy over a broad range of dataset types, yet do so with comparative computational efficiency. Random Forest has been further enhanced by parallelization, allowing the process of constructing decision trees to occur in parallel on systems with multi-core processor units. As a result, multi-threaded versions of the algorithm have enabled greater processing speeds \cite{RN533}.\\
\indent With the rise of smartphones and Internet of Things (IoT) over recent years, there is growing potential to bring data mining to hardware-constrained devices, despite the reduced levels of processing speed they may offer. As an example, our previous research \cite{RN598} has brought locally-executed data mining to the Android platform through our open-source DataLearner app, available free on Google Play\footnote{https://play.google.com/store/apps/details?id=au.com.darrenyates.datalearner}. Moreover, moves by cloud service providers such as Amazon Web Services to change the cost structure of these services from a per-hour to a per-second basis \cite{RN496} suggests there can also be direct cost savings by implementing more efficient algorithms in cloud-based applications. Thus, the need continues for further research into algorithms that deliver faster performance without sacrificing accuracy.\\
\indent FastForest is an optimised derivative of Random Forest that achieves improved levels of processing speed whilst maintaining (and frequently exceeding) the classification accuracy of Random Forest. This is achieved through the implementation of three components detailed in Section 3 - Subsample Aggregating (or ‘subbagging’), Logarithmic Split-Point Sampling (LSPS) and Dynamic Restricted Subspacing (DRS). 

\begin{table}
\centering
\caption{Notation used in this paper}
\label{tab:1}       
\begin{tabular}{llll}
\noalign{\smallskip}\hline\noalign{\smallskip}
DRS & Dynamic Restricted Subspacing & $k$ & Size of attribute subspace \\
LSPS&Logarithmic Split-Point Sampling	& $M$ &Attribute space of $D$\\
$A$	&Attribute candidate at tree node	&$n$&	Number of records in $D$\\
$a$	&Subbagging factor&	$p_i$	&Prob. record in $D$ has class value $c_i$\\
$c$	&Number of distinct class values &	$r$	&Random subsample on M\\
$D$	&Initial training dataset&	$s$&	Number of split-point candidates\\
$D_i$&Records in $D$ at tree node $i$	&$T$&	Ensemble of decision trees, size $|T|$\\
$D_j$&$j^th$ bootstrap sample of $D$		\\

\noalign{\smallskip}\hline
\end{tabular}
\end{table}

\subsection{Original contributions}
As will be detailed throughout this paper, the original contributions of this research include:\\
\begin{itemize}
\item Empirical testing of subbagging scale factors from 0.05 to 0.632 within Random Forest over 30 freely-available numerical and categorical datasets (Section 3.1).
\item	Development of the Logarithmic Split-Point Sampling (LSPS) component, inspired by our previous work on split-point sampling in the single-tree SPAARC algorithm \cite{RN461} (Section 3.2).
\item Development of the Dynamic Restricted Subspacing (DRS) component, inspired by ‘dynamic subspacing’ \cite{RN451} to achieve greater speed without loss of accuracy (Section 3.3).
\item Combining Subbagging, Logarithmic Split-Point Sampling (LSPS) and Dynamic Restricted Subspacing (DRS) components into an ensemble classifier called `FastForest' that appears to be faster than Random Forest without sacrificing accuracy.
\item Broad-scale empirical testing of FastForest over a cache of 45 categorical, numerical and mixed freely-available datasets on PC and smartphone hardware (Sections 4.1 to 4.5).
\end{itemize}
Common notation used in this paper is shown in Table 1. This paper continues with Section 2 covering the basics of ensemble classifiers and related efforts in Random Forest optimisation, Section 3 details the components of FastForest, while Section 4 covers experimentation and comparison with various ensemble algorithms, including Random Forest itself. Section 5 concludes this paper.
\section{Related Work}
\label{sec:2}
Ensemble classifiers, such as Random Forest, are classification systems that combine the output of multiple individual learning models or `classifiers' to more accurately categorise or `classify' not only training datasets, but also new previously-unseen dataset records. By combining the outputs of multiple individual models, ensemble classifiers are often more robust to issues such as errors or ‘noise’ in dataset record values \cite{RN538}. They are also often better able to identify multiple patterns within the data than a single classifier \cite{RN537} and as such, have long been an area of research \cite{RN455,RN454,RN468,RN497,RN536}. 
\subsection{The `decision tree' base classifier}
The individual or `base' learning algorithm used by many ensemble classifiers is the decision tree \cite{RN539}, an example of which is shown in Fig. 1a. It is a flowchart or tree-like structure consisting of internal branching points or `nodes' and terminating nodes or `leaves' that feature the designated class value or `label'. The example tree in Fig. 1a is built from the training dataset in Fig. 1b. This simple dataset consists of three attributes (columns) and six records (rows). The designated ‘class’ attribute in this dataset is the `Mortgage' attribute. The goal would be to identify a data pattern relating age and salary to having a mortgage.

\begin{figure}
\centering
\includegraphics[scale=0.22]{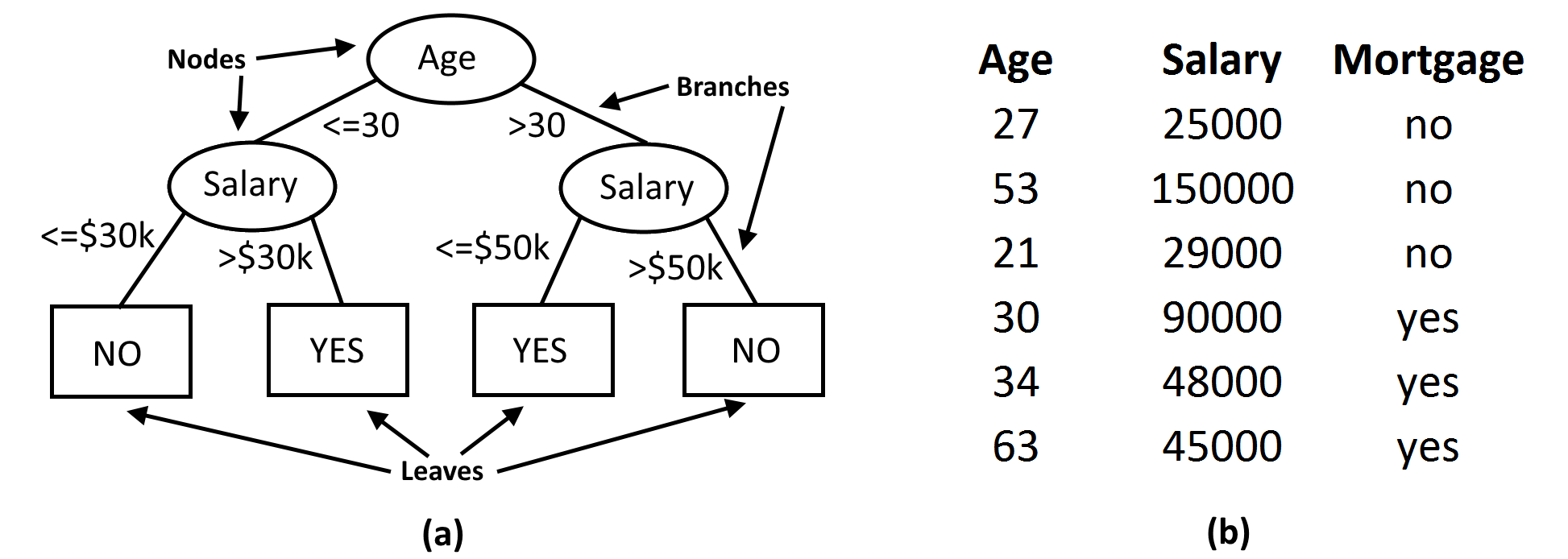}
\caption{The decision tree in (a) consists of nodes (ovals), branches (edges) and terminating nodes or `leaves' (rectangles). This tree correctly classifies all six records in dataset (b) by their Mortgage attribute values.}
\end{figure}

\indent The tree building or `induction' process begins by finding the initial non-class attribute and its value that best splits the records by their `Mortgage' attribute values. An attribute's ability to split records at a node can be estimated by its `information gain'. Multiple methods for calculating information gain exist, including the Gini Index used by the CART algorithm \cite{RN462}, Gain Ratio featured in C4.5 \cite{RN463} and the original `Information Gain' method for the earlier ID3 decision tree \cite{RN549}. Thus, the dataset attribute that achieves the greatest information gain at a particular node is selected to become the splitting attribute at that node. In addition, the attribute value at which the maximum information gain is achieved also becomes the `split point' for records at that node. \\
The process can be demonstrated mathematically using the equation for information value or `entropy' developed by Shannon \cite{RN539}:
\begin{equation}
Info(D) = - \sum\limits_{i=1}^{c} p_ilog_2(p_i)
\end{equation}
where $D$ is the dataset, $c$ is the number of distinct class values (two, in this example - $‘yes’$ and $‘no’$) and $p_i$  is the probability of a record in $D$ belonging to a particular class value $c_i$. In this example, the dataset in Fig. 1b features three records with $‘yes’$ Mortgage value and three with $‘no’$. Thus, the uncertainty of the dataset can be calculated using (1):
\begin{equation}
Info(D) = - (\frac{3}{6}log_2(\frac{3}{6}) + \frac{3}{6}log_2(\frac{3}{6})) = -(-0.5+(-0.5)) = 1.0bits\nonumber\\
\end{equation}
The ‘Information Gain’ is then the improvement or ‘gain’ in information achieved by splitting the records against one of the non-class attributes \cite{RN539}. This gain is calculated by:

\begin{equation}
Gain(D)=Info(D)-Info_A(D) 			
\end{equation}
where $\mathrm{Info}_\mathrm{A}(D)$ is the information value required to determine a record's class by splitting on attribute $A$. In a decision tree build process, each non-class attribute of the dataset is tested in turn as attribute $A$ at each node. This is done by splitting the records by each candidate attribute value one-by-one and calculating the information gain with equations (1) and (2). The aim is to identify the attribute-value pair that delivers the maximum information gain. 
Fig. 2 extends the example of Fig. 1 to show the process for the first two attribute value splits. Starting with the $‘Age’$ attribute as the test candidate, Fig. 2a shows the initial division of records based on the $‘Age’$ attribute and the value of $25$. The information value can be calculated via Equation (1):
\begin{equation}
Info_A(D)=\frac{1}{6}\times(-\frac{0}{1}log_2\frac{0}{1}-\frac{1}{1}log_2\frac{1}{1}) + \frac{5}{6}\times(-\frac{3}{5}log_2\frac{3}{5}-\frac{2}{5}log_2\frac{2}{5}) = 0.809bits\nonumber\\
\end{equation}
Fig. 2b shows how the $‘Age’$ attribute and test split-value of $25$ divides the dataset records into two groups. The group consisting of $Age$ values less than or equal to ($\leq$) $25$ has the one record, with $Mortgage$ value of $yes$. The group with $Age$ values greater than ($>$) $25$ has five records in all - three records with $Mortgage$ value of $no$ and two with $yes$. This delivers the information value of 0.809 bits calculated above. The `information gain' achieved by this split-value is calculated using equation (2):
\begin{equation}
Gain(D)=Info(D)-Info_A(D)=1.0-0.809=0.191 bits\nonumber\\
\end{equation}
Thus, this split-point provides some information gain, but to identify the value delivering the greatest information gain, each attribute value must be tested. The next split value of ‘27’, shown in Figs. 2a and 2c, has information value:
\begin{equation}
Info_A(D)=\frac{2}{6}\times(-\frac{1}{2}log_2\frac{1}{2}-\frac{1}{2}log_2\frac{1}{2})+\frac{4}{6}\times(-\frac{2}{4}log_2\frac{2}{4}-\frac{2}{4}log_2\frac{2}{4})=1.0 bit\nonumber
\end{equation}
Again, the information gain for this attribute value comes via equation (2) as 1.0 - 1.0 = 0. Thus, this split-value provides no information gain and can be disregarded as a potential candidate, since it is less than the 0.191-bit information gain achieved using the value `25'. This can be also seen intuitively from the split of $Mortgage$ attribute values, shown in Fig. 2c.  Drawing a line under the second record in Fig. 2a  and splitting on $Age$ $\leq$ 27 and $Age$ $>$ 27 creates the two groups as shown in Fig. 2c - each group features equal numbers of $yes$ and $no$ records and is of no help in classifying a new record.

\begin{figure}
\centering
\includegraphics[scale=0.18]{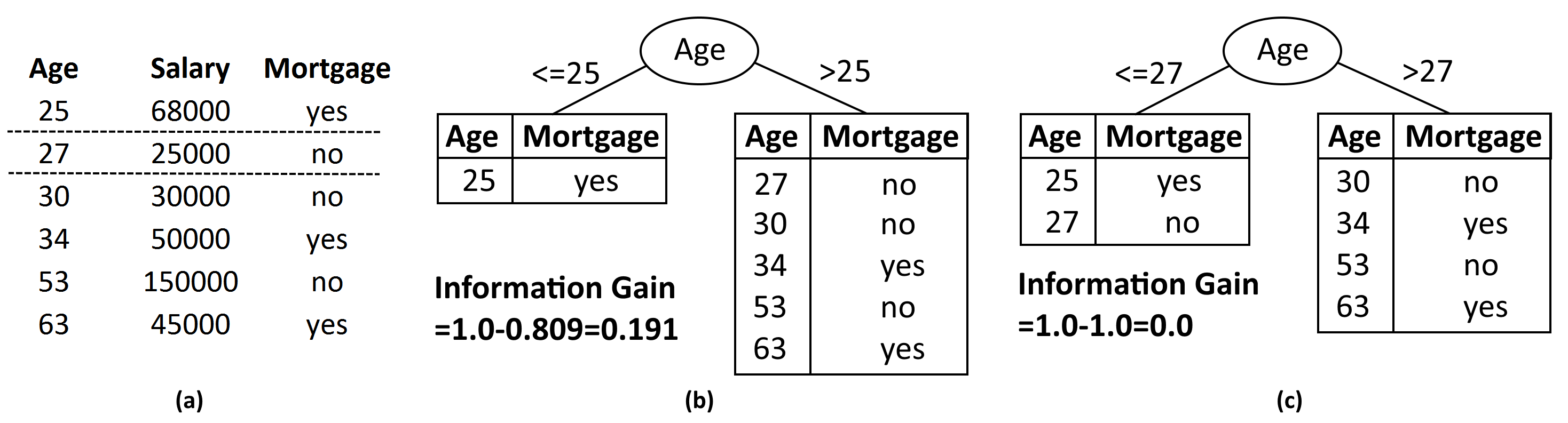}
\caption{The attribute splitting process involves (a) splitting the records into two groups by each attribute value in turn, with (b) showing the split on an attribute value of `25' and (c) an attribute value of `27'.}
\end{figure}

\indent Following this process through all of the $Age$ attribute values, the highest information gain is achieved with a split-value of `30', where the gain is 1.0 - 0.459 = 0.541 bits. However, testing the $Salary$ attribute shows the split-value of `45000' also provides a maximum information gain of 0.541 bits. In this situation where two attributes achieve equal maximum information gain, one is chosen at random, thus $Age$ can be selected and becomes the root node attribute (top oval in Fig. 1a). This testing process is used to create further nodes for each split of dataset records and continues recursively until an end-point or `termination' criterion is met. End-point criteria include a node reaching a user-defined level of maximum tree depth, too few records reaching a node, as well as no further division of records possible \cite{RN539}. \\
\indent However, identifying $Age$ as the root node attribute and `30' as its split value required testing all non-class attributes and all distinct attribute values - in this example, 12 tests in all. Moreover, this process must be repeated for every node in the tree, and then, for every tree in the ensemble classifier.\\
\indent Once the tree is complete, an unlabelled record (one without a class value) can be analysed by the tree. The new record's non-class attribute values determine the branches taken until one of the leaf nodes is reached. If the class values of all records at the selected leaf are the same, this value becomes the class value of the unlabelled record. If, however, the leaf consists of multiple class values, some method for selecting between the class values must be chosen. It is common for decision trees to simply choose the class value that is most frequent (known as the `majority' class).
\subsection{Ensemble classifiers}
Ensemble classifiers are numerous, but few are more well-known than ‘Bootstrapped Aggregating’, commonly known as `Bagging' \cite{RN455}. This method generates multiple copies of an original dataset $D$ (size $|D|$ = $n$) through a process of ‘bootstrapping’. This involves choosing $n$ records sampled at random from dataset $D$ with replacement, that is, any sample chosen for a bootstrap may be chosen again for the same bootstrap. An example using the original dataset from Fig. 1 is shown in Fig. 3. One record from the dataset in Fig. 3a is chosen at random, recorded, put back and the process completed six times each to create the datasets in Figs. 3b and 3c.

\begin{figure}
\centering
\includegraphics[scale=0.3]{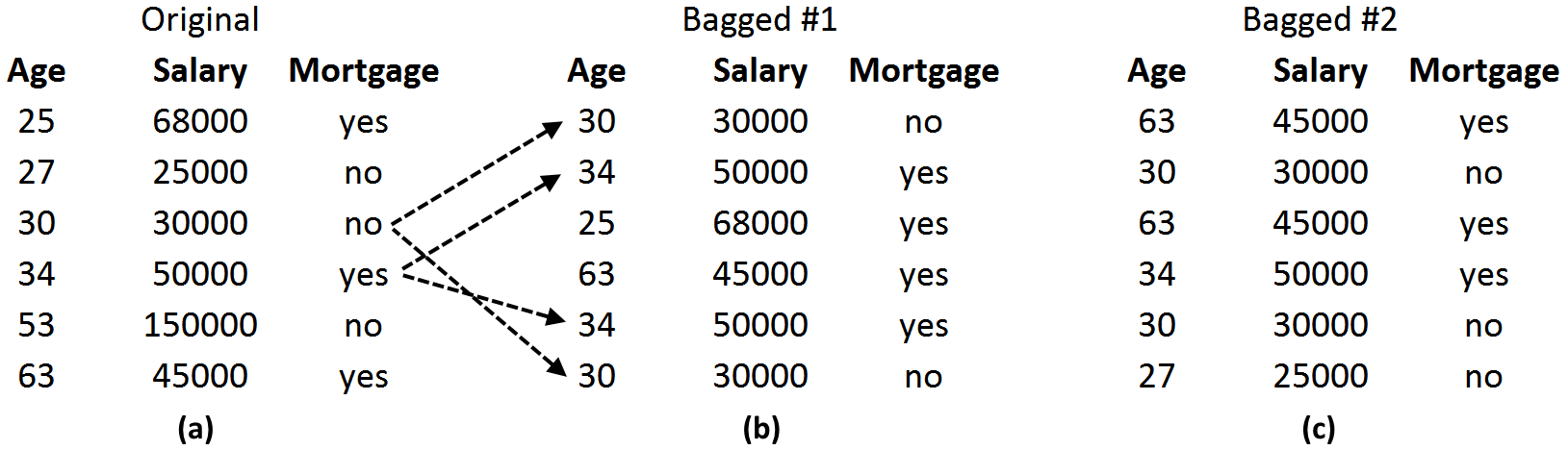}
\caption{The original Fig. 1 dataset (a) resampled with replacement to create 'bagged' dataset samples (b) and (c).}
\end{figure}

\indent The purpose of Bagging is to achieve a set of more diverse dataset derivatives. However, this `with replacement' technique results in only approximately 63.2\% of the records selected being unique, with the remainder as duplicates \cite{RN498}. Each new sample dataset $D_j$ is then used to induce a new decision tree, with as many datasets created as required, such that $j$=[1,$|T|$], where $T$ is the ensemble of trees, size $|T|$. The results of each decision tree are combined or `aggregated', either through averaging for a regression problem, or through a voting method (such as majority voting) for a classification task. Thus, the bootstrap aggregation creates an ensemble of decision trees $T$, where the only variation is the dataset $D_j$ used to induce each tree $T_j$.\\
\indent An alternate sampling method is `subsample aggregation' or `subbagging' \cite{RN470}. It is similar to Bagging in that record sampling is used to create $|T|$ bootstrap datasets as before, however, records in each dataset are chosen at random without replacement - that is, each record can only be sampled once in each new bootstrap dataset $D_j$, but may appear in multiple datasets. However, since sampling without replacement $n$ times simply regenerates the original dataset $D$, only a portion of $D$ is sampled in subbagging. A method proposed in previous research \cite{RN470,RN524} is $|D_j|$ = $an$, with 0\textless$a$\textless 1. Again, using Fig. 3a as the example, three distinct records would be selected to create dataset $D_j$ if $a$ = 0.5. \\
\indent Friedman and Hall \cite{RN524} identified through simulation that bagging and `half-subbagging', where $a$ = 0.5, are said to be ‘almost identical’ in terms of the overall mean-squared error (MSE) when producing single-node binary trees with two terminal nodes or ‘leaves’. This was shown quantitatively by Buhlmann and Yu \cite{RN470}. Moreover, Buhlmann and Yu noted that subbagging offers improvements to variance whilst adding only slightly to the bias, factors also seen in Bagging (ideally, there should be maximum variance between the dataset copies and minimum bias). In addition, the increased bias introduced by subbagging decreases as $a$ increases. As a result, potential exists for `half-subbagging' to offer comparable performance to traditional Bagging in overall classification accuracy. Further, the reduction in the number of records processed through half-subbagging also reduces the computational cost and improves processing speed.\\
\indent A study by Zaman and Hirose \cite{RN499} examined subbagging scale factors, comparing them with the standard bagging approach, but featured only a limited number of datasets. Moreover, it used `stable' classifiers, including Linear Support Vector Machine (LSVM) and Stable Linear Discriminant Analysis (SLDA). By contrast, Random Forest, like other decision tree-based algorithms, is considered to be `unstable', since a small change in one node can drastically alter the tree structure. However, in spite of this instability, Buhlmann \cite{RN525} suggests half-subbagging is a `good choice', yielding `very similar empirical results' to bagging when used in decision trees and further noted that half-subbagging can save `more than half the computing time' compared with standard bagging.\\
\indent Another popular ensemble technique is Random Subspacing \cite{RN456}. Instead of using the full attribute space $M$ to test at each node during the building of a decision tree, this technique instead employs a random subsample of attributes $k$ for the test, such that $k$ $\ll$ $M$. Each tree in the ensemble is built using a different randomly-selected subsample of attributes. Unlike Bagging/Subbagging, all trees are generated from the original dataset $D$. However, as with Bagging and Subbagging, Random Subspace supports parallel learning - this means faster processing speed. Random Subspacing has also been shown to greatly improve classification accuracy over Bagging \cite{RN456}. Moreover, testing \cite{RN456} revealed that an attribute subspace size of 0.5 (50\%) was shown to produce best results on 14 test datasets.

\subsection{Random Forest}
Following this work \cite{RN456}, Breiman later combined Bagging with Random Subspacing, creating the `Random Forest' algorithm \cite{RN454}. According to Breiman, the random selection of records and attributes delivered good results for classification (categorical class attribute) tasks, but was less successful initially in regression (numerical class attribute) problems. Nevertheless, Random Forest is robust against errors or `noise' in a dataset, does not overfit the training dataset (meaning it can generalise well with similar unseen records) and with a sufficient number of trees, the overall classifier's prediction or `generalisation' error rate converges to a limit. In addition, Random Forest is considered a fast classifier \cite{RN538}, but more recently, parallelised versions can take advantage of the multi-core processors now available. For example, the implementation of Random Forest available in Weka release version 3.8 \cite{RN533,RN599} enables one tree per processor core to be induced concurrently, reducing the forest build time. Weka is the open-source graphical user interface (GUI)-based data-mining application developed by the University of Waikato.\\
\indent However, parallelisation presents challenges for some techniques, with Boosting \cite{RN525} an example of this. Boosting is a sequentially-built or `dependent' classifier framework that builds individual trees successively using corrective factors carried over from the preceding tree. Parallelising sequential frameworks, such as Boosting and ForestPA \cite{RN468} for example, is a considerably more complex task. \\
\indent Another area of research into improving Random Forest's computational efficiency has been to identify the optimum number of trees developed within the forest \cite{RN452,RN453}. Since the complexity of Random Forest is directly related to the number of trees, it is logical that reducing the tree count would also reduce the overall computational load. This has been previously researched \cite{RN452}, with the McNemar Test for Significance used to identify a threshold of optimality after each tree is sequentially developed. However, again, the sequential nature of this method is at odds with the parallelisation now available in Random Forest implementations. In addition, research undertaken by Oshiro, Perez, and Baranauskas \cite{RN453} found over a test of 29 datasets that there was no significant difference in overall classification accuracy between using a particular number of Random Trees and its double, for example between $|T|$=128 and $|T|$=256. However, this research did find that once the tree count reached 128, there was no significant change in accuracy with any tree counts in binary increments up to 4,096 trees \cite{RN453}. 
\subsection{Speed optimisations beyond parallel processing}
Thus, to take advantage of parallelisation, optimisations originating within the base decision tree itself appear to offer the best potential. Previous examples of in-tree optimisations include the SPAARC decision tree algorithm \cite{RN461} that incorporates a technique called `Split-Point Sampling' (SPS). This single-tree algorithm follows standard tree induction practices by testing attributes at each node to determine which one creates the most homogenous split of records. However, the Split Point Sampling (SPS) technique presented in SPAARC \cite{RN461} reduces the number of potential split-point candidates for numerical attribute values to a maximum count of 20, equi-distant across the attribute value range. It was shown empirically that SPS reduces processing time without significantly affecting the overall classification accuracy of the decision tree. However, this testing did not include how SPS responds in an ensemble framework.\\
\indent Nonetheless, while improving ensemble processing speed is the key focus of this work, it should not come at the expense of classification accuracy. Previous research \cite{RN451} identified that as a dataset is progressively partitioned during tree induction, fewer attributes are likely to be relevant to the fewer records that reach the lower (deeper) segments of the tree. The `dynamic subspacing' solution \cite{RN451} progressively increases the size of the attribute candidate subspace, $k$, to encourage more accurate record splitting. A bonus of this technique is that it works on each tree independently, allowing for parallelisation, and limits the increase in attribute candidates (and thus, computational load) to when the record counts are least. \\
\indent In addition, there are algorithm-acceleration methods based on hardware for improving processing speed, such as general-purpose computation on graphics processor unit (GPGPU) technology \cite{RN535}. However, as powerful as these solutions can be, they do require specialised hardware (GPU-based PC video cards for example) and may not suit everyday users locally executing data-mining on standard PCs or smartphones. Thus, techniques to further improve the two seemingly opposing factors of classification accuracy and processing speed should ideally be executable on standard PC (or smartphone) hardware and work on existing software platforms. It is research in this area that this paper now investigates.

\section{Our `FastForest' Algorithm Framework}
This section will now introduce the components used in our proposed `FastForest' algorithm, starting with Subbagging (Section 3.1), followed by Logarithmic Split-Point Sampling (3.2), Dynamic Restricted Subspacing (3.3) and conclude with algorithms for the proposed components in Section 3.4.
\subsection{Subbagging}
As noted in Section 2.2, Bagging involves choosing $n$ samples from the $n$ records within dataset $D$ ($|D|$ = $n$), each time with replacement, so that any selected sample may be chosen again. This `bootstrap' process enables any number of diverse dataset derivatives to be generated. However, with typically only 63.2\% of selected records being unique due to the `with replacement' selection, the remaining records in a bagged dataset can represent unnecessary work for the algorithm to perform. This can reduce processing speed. Moreover, analysis of the Weka 3.8.3 implementation of Random Forest suggests that the algorithm introduces a weighting factor to allow duplicate records to be removed from each bagged dataset after it is generated, leaving only approximately 0.632 $\times$ $n$ distinct records in each dataset $D_j$. This reduces the processing workload during construction of each tree. 

\begin{figure}
\centering
\includegraphics[scale=0.3]{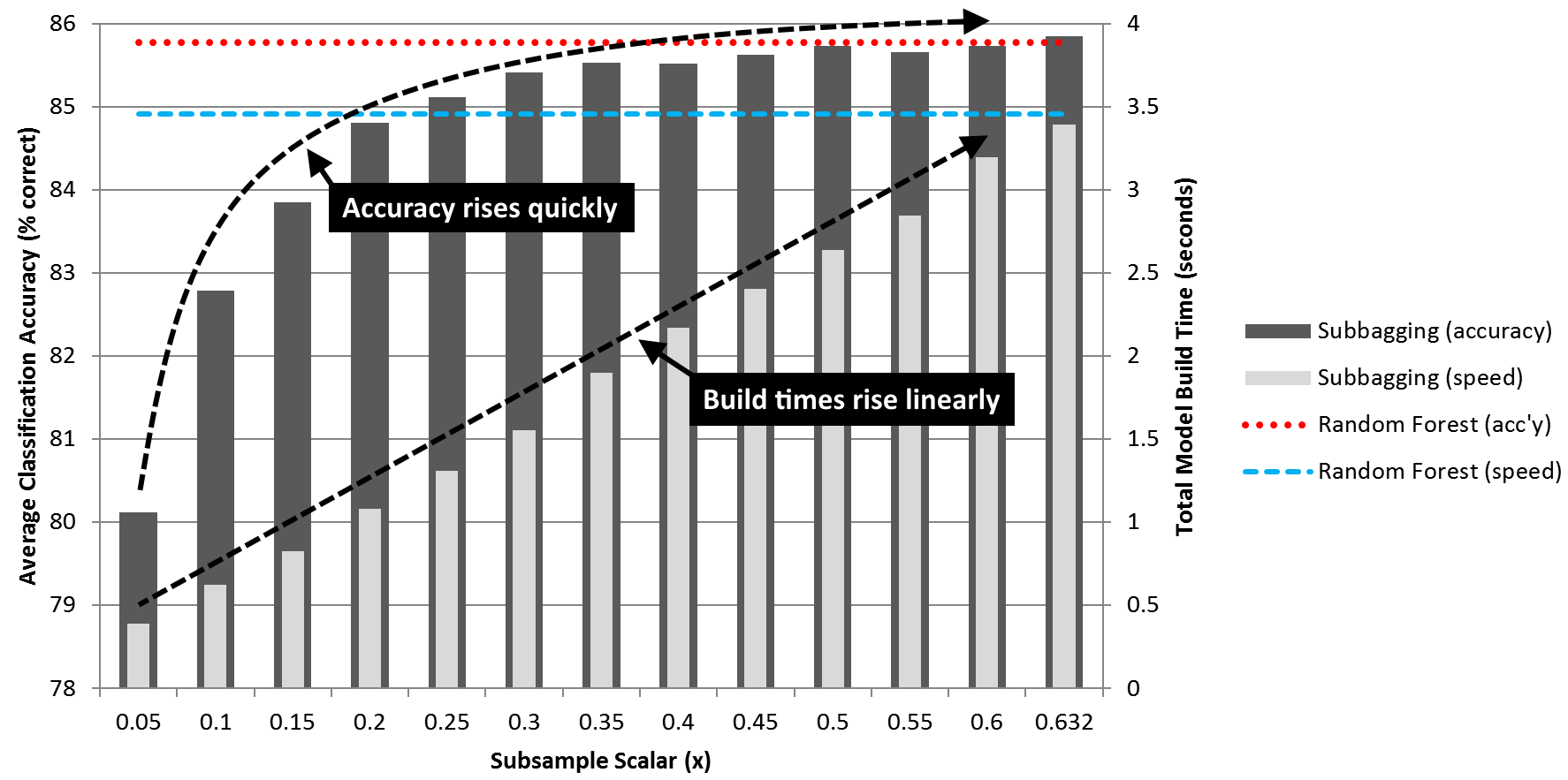}
\caption{Total build speed and average classification accuracy of subbagging on 30 datasets of Groups 1 and 2 detailed in Tables 2 and 3 compared with Random Forest (RF, 85.77\% accuracy / 3.45 secs speed).}
\end{figure}

\indent The proposed subbagging component of our FastForest algorithm, by contrast, only selects unique records, but the question now becomes how many records do you select? At the two extremes, selecting too few records limits individual tree accuracy (at its limit, setting the size of each dataset $D_j$ to $|D_j|$=0), but selecting too many ($|D_j|$=$|D|$) removes bag diversity. Empirical testing over 30 training datasets shows that a subbagging factor, $a$, of at least 0.3, such that $|D_j|$ = $an$, delivers an average accuracy to within 0.5\% of Random Forest. The results are shown in Fig. 4.\\
\indent Two versions of Random Forest were used to conduct these tests. The first was the standard Random Forest implementation found in the Weka version 3.8.3 software. The second was a modified version of the same Random Forest algorithm but with the standard Bagging component replaced by Subbagging. The two horizontal lines in Fig. 4 represent the average accuracy (85.77\%, upper dotted line) and total build time (3.45 seconds, lower dashed line) across 30 datasets detailed in Tables 2 and 3 of Section 4 for the standard `Bagging' version of Random Forest. The darker-shaded columns represent the average classification accuracy of the modified `Subbagged' version of Random Forest (referred to as `SB') across the same datasets over a range of subbagging scale factors from 0.05 to 0.6 in 0.05-step intervals (the 0.632 factor is to emulate Bagging). The lighter-shaded columns show total build speed over the same datasets and subbagging factors for this modified algorithm. Thus, ideally, the darker columns should exceed the `Random Forest average accuracy' line (the upper dotted line in Fig. 4) and the lighter columns remain below the `Random Forest build time' line (the lower dashed line in Fig. 4).\\
\indent What is noticeable is that the average classification accuracy rises sharply with comparatively low sub-bagging factors, to be well within 0.5\% of RF by the time the subbagging scale factor reaches just 0.3. However, at the same time, the growth in total build time is essentially linear. At the half-subbagging ($a$ = 0.5) level, the accuracy of SB is less than 0.05\% below RF, while still achieving a 23\% speed gain over RF.\\
\indent Moreover, it was found (through results presented later in this paper) that half-subbagging not only delivers substantial processing speed gains over RF, but in combination with Logarithmic Split-Point Sampling (LSPS) and Dynamic Restricted Subspacing (DRS), can match and frequently exceed RF for classification accuracy. As a result, the FastForest algorithm uses `half-subbagging', such that:
\begin{equation}
|D_j| = 0.5 \times n
\end{equation}
This ensures each dataset subsample $D_j$ consists of 0.5 $\times$ $n$ distinct randomly-selected records, reducing the induction time for each tree and for the forest as a whole, yet with little effect on classification accuracy.

\subsection{Logarithmic Split-Point Sampling (LSPS)}
As discussed in Section 2.1, the process of tree induction involves testing attributes at each node to determine the attribute and attribute value that best splits the records at the node into the most distinct groups. The choice of possible split points for categorical attributes, where attributes have a finite value space, is straightforward - each distinct attribute value becomes a split-point.

\begin{figure}
\centering
\includegraphics[scale=0.5]{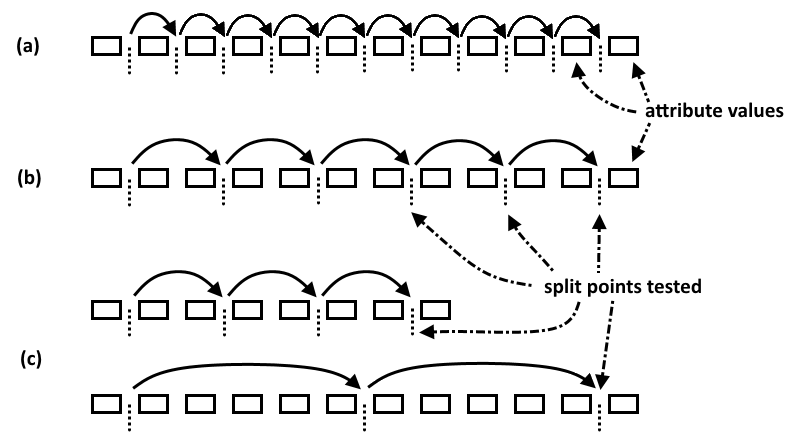}
\caption{Split point candidates selected in (a) Random Forest between each adjacent pair, (b) SPAARC using a fixed count of 20 candidates, and (c) Logarithmic Split-Point Sampling (LSPS) using log2 of the record count at each node. }
\end{figure}

\indent However, for numerical attributes, Random Forest employs the binary split method shown in Section 2.1, so that records with attribute value less than or equal to the split value are split into one branch; those with values greater than the split value are split into another branch. The question then becomes which split value or `split-point' provides the largest information gain and how do you go about finding it?\\
\indent Fig. 5 outlines three different techniques used by (a) Random Forest, (b) our previous SPAARC algorithm and (c) the FastForest algorithm presented in this paper. In each diagram, the row of rectangles represents the ordered record values of the candidate attribute, the dotted vertical lines between values are the tested mid-points of adjacent values and the solid arrows represent the method of selecting the split-points to be tested. Fig. 5a represents an implementation of Random Forest that tests every adjacent pair of record values. While this does eventually find the optimum split-point value, the task of testing every adjacent value pair consumes processing time.\\
\indent To compensate for this, our previous SPAARC algorithm \cite{RN461} incorporated a simplified split-point sampling technique shown in Fig. 5b whereby a maximum of 20 split-points were tested equidistance across the attribute value range. If the number of record values at the node $i$, denoted as $D_i$, was less than 20, the number of split-points was reduced to $D_i$-1. Overall, this created a `hopping' effect, as candidate split-points were skipped to reduce the number of split-points tested. However, while this was shown to work empirically in a single-tree environment such as SPAARC, the choice of `20' candidate split-points was arbitrary.\\
\indent In comparison, the proposed Logarithmic Split Point Sampling (LSPS) component of FastForest in Fig. 5c ensures that for a large number of records, the number of candidate split-points remains low. However, as the record count reduces, such as when the tree induction process nears a leaf, the number of possible candidate split-points proportionally increases.\\
\indent Thus, the purpose of the LSPS component of FastForest is to reduce the number of candidate split points (and thus, the number of tests required) to $s$, where $s$ $\ll$ $D_i$. This is done using the equation:
\begin{equation}
s=int(log_2(D_i ))+1
\end{equation}
where $D_i$ is the number of records with non-missing attribute values reaching the node $i$.
An ‘A-B’ test of this improvement to reduce processing time showed no ill effect on classification accuracy. With the FastForest algorithm alternately set between $s$ = 20 (test A) and $s = log_2(D_i)+1$ (test B) over the Group 1 and Group 2 datasets detailed in Section 4, test B completed the build process in 2.83 seconds, ahead of test A's 3.04 seconds, for a gain of 6.9\%. At the same time, test B also achieved a slightly higher overall classification accuracy of 86.03\% versus 85.95\%. In terms of comparative wins, there was essentially no difference, with Test B achieving 11 wins, 10 ties and 9 losses. Thus, it appears speed is gained without loss of accuracy.

\subsection{Dynamic Restricted Subspacing (DRS)}
The purpose of Dynamic Restricted Subspacing (DRS) is to deliver improvements in classification accuracy and overcome any accuracy losses that may result from subbagging or LSPS, but without overly compromising the speed gains they achieve. Evidence of this will be shown in Section 4.3.3.\\
\indent If we determine a training dataset $D$ has attribute space $M$, such that the attribute space has size $|M|$, the original Random Subspacing method built into Random Forest selects a subset or `subspace' $k$ of $M$, from which attributes can be drawn for testing at each tree node. Methods for determining the size of $k$ put forward in literature, including 1 and $sqrt(M)$ \cite{RN530}. A common option (and implemented within the Weka 3.8.3 version of RandomForest) is the method:
\begin{equation}
k = int(log_2 (|M|))  + 1
\end{equation}
However, research by Adnan and Islam \cite{RN451} argued that as the training dataset is split or `partitioned' via each tree node, the number of attributes that relate to the class attribute falls, reducing the likelihood of relevant attributes being selected deeper down the tree nodes (that is, the further away from the root node). As a result, it is possible for this selection of lesser-quality attributes to result in lesser-quality splits, which may reduce the accuracy of individual trees and the ensemble/forest as a whole. The authors proposed `Dynamic Subspacing' to increase the attribute subspace size, $k$, in response to the falling number of records at a particular tree node. Thus, with a training dataset $D$ of size $|D|$ and the number of records reaching the particular node $i$ as $|D_i|$, the attribute subspace, $k$, is determined by:
\begin{equation}
k = int(log_2 (|M|\times\frac{|D|}{|D_i |}))  + 1	
\end{equation}
This `Dynamic Subspacing' method increases the attribute subspace as the number of records reaching a node falls. It can also improve classification accuracy, particularly for datasets where good attributes are few and/or dimensionality is high \cite{RN451}. \\
\indent Dynamic Subspacing also offers a secondary benefit - it limits the size of the random subspacing nearer the root of the tree, where records at each node are more plentiful than nearer each leaf. This helps maintain processing speed. For example, at the root node, where the number of records $D_i$ at the node is the full dataset $D$, the $D/D_i$ factor of equation (6) reduces to `1' and the equation itself reverts to its original `Random Subspacing' form in equation (5). Thus, if the full attribute space is $M$=100, the selected subspace becomes $k$=7. However, if the number of records at test node $D_i$ falls to 1/100th of $|D|$ (that is, $D$/100), equation (6) now increases the number of attributes to $k$=14. As a result, while this increase in the attribute subspace helps promote selection of a high-quality attribute, the extra processing workload required is somewhat offset by the lower number of records now reaching the test node.

\begin{figure}
\centering
\includegraphics[scale=0.3]{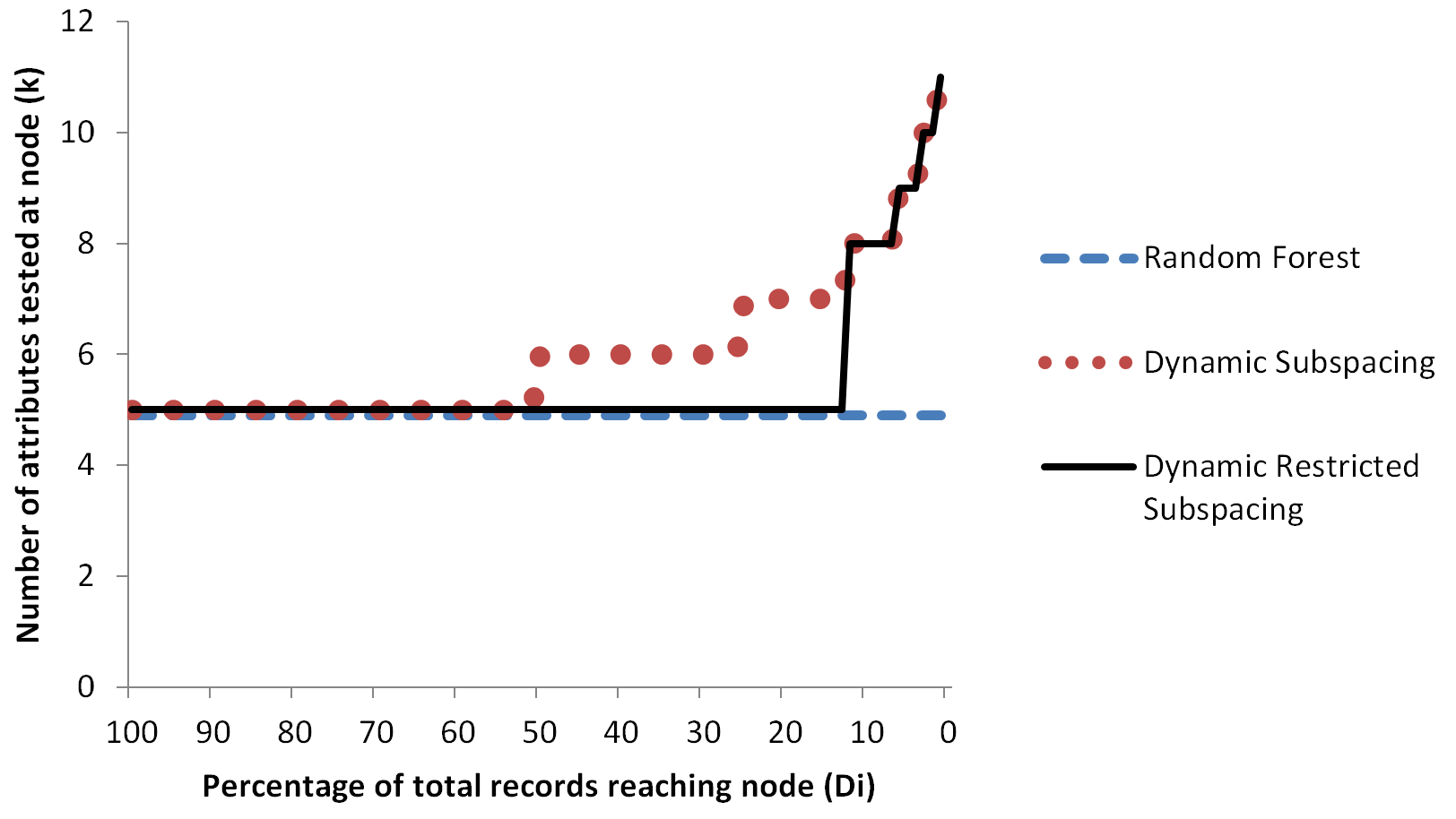}
\caption{Example showing Dynamic Restricted Subspacing (DRS) increasing the number of node attributes, $k$, only once the number of records at the node falls below 12.5\% (1/8) of the total record count.}
\end{figure}

\indent However, from a processing perspective, the addition of new attributes to the subspace $k$ increases the processing workload at each node. Fig. 6 shows the progression of $k$ against the falling proportion of records reaching the test node, assuming a hypothetical dataset with an attribute count, $|M|$ = 16. Random Forest reduces the attribute space to a smaller fixed subspace of $k$=5, represented in the Fig. 6 graph by the horizontal dashed line. In contrast, the previously-proposed Dynamic Subspacing method, shown as the dotted line in Fig. 6, begins increasing the size of the attribute subspace even when the proportion of records reaching the node is still 50\%. Thus, the size of $k$ increases by one for every halving in the number of records reaching a node.\\
\indent This raises the question of whether it is necessary to increase the attribute subspace so early (as with Dynamic Subspacing), while a high proportion of records still appear at a node. Moreover, if 50\% is deemed too high a threshold, at what proportion of records remaining should $k$ begin to increase?\\
\indent Our proposed Dynamic Restricted Subspacing (DRS) limits the application of equation (6) until the number of records reaching a node falls to one-eighth (12.5\%) of the total number of records in the dataset, at which time, it is applied to boost the size of $k$. The area in Fig. 6 where the previous Dynamic Subspacing method and our proposed Dynamic Restricted Subspacing (DRS) component differ is where speed gains are achieved by DRS. \\
\indent The choice of `threshold' factor is practically limited to between $|D|/2$, in which case, it is the same as the previous Dynamic Subspacing method, and zero, which is Random Forest. Thus, $|D|/8$ is a compromise between the two, enabling speed gains over the previous method, but also ensuring Dynamic Restricted Subspacing (DRS) is applied early enough to still achieve accuracy gains. Thus, DRS sets $k$ as follows:\\
\begin{equation}
k = \left\{
  \begin{array}{lr}
    $equation (5)$ & : |D_i| > |D|/8\\
    $equation (6)$ & : |D_i| \le |D|/8
  \end{array}
\right.
\end{equation}
Moreover, tests between the two techniques in Section 4.3.4 over the 45 datasets of groups 1, 2 and 3 indicate that not only does Dynamic Restricted Subspacing help further reduce the tree induction time by more than 6\%, it has no apparent negative effect on classification accuracy. Finally, this combination of half-subbagging, Logarithmic Split-Point Sampling (LSPS) and Dynamic Restricted Subspacing (DRS) components creates the FastForest algorithm.

\begin{figure}
\centering
\includegraphics[scale=0.5]{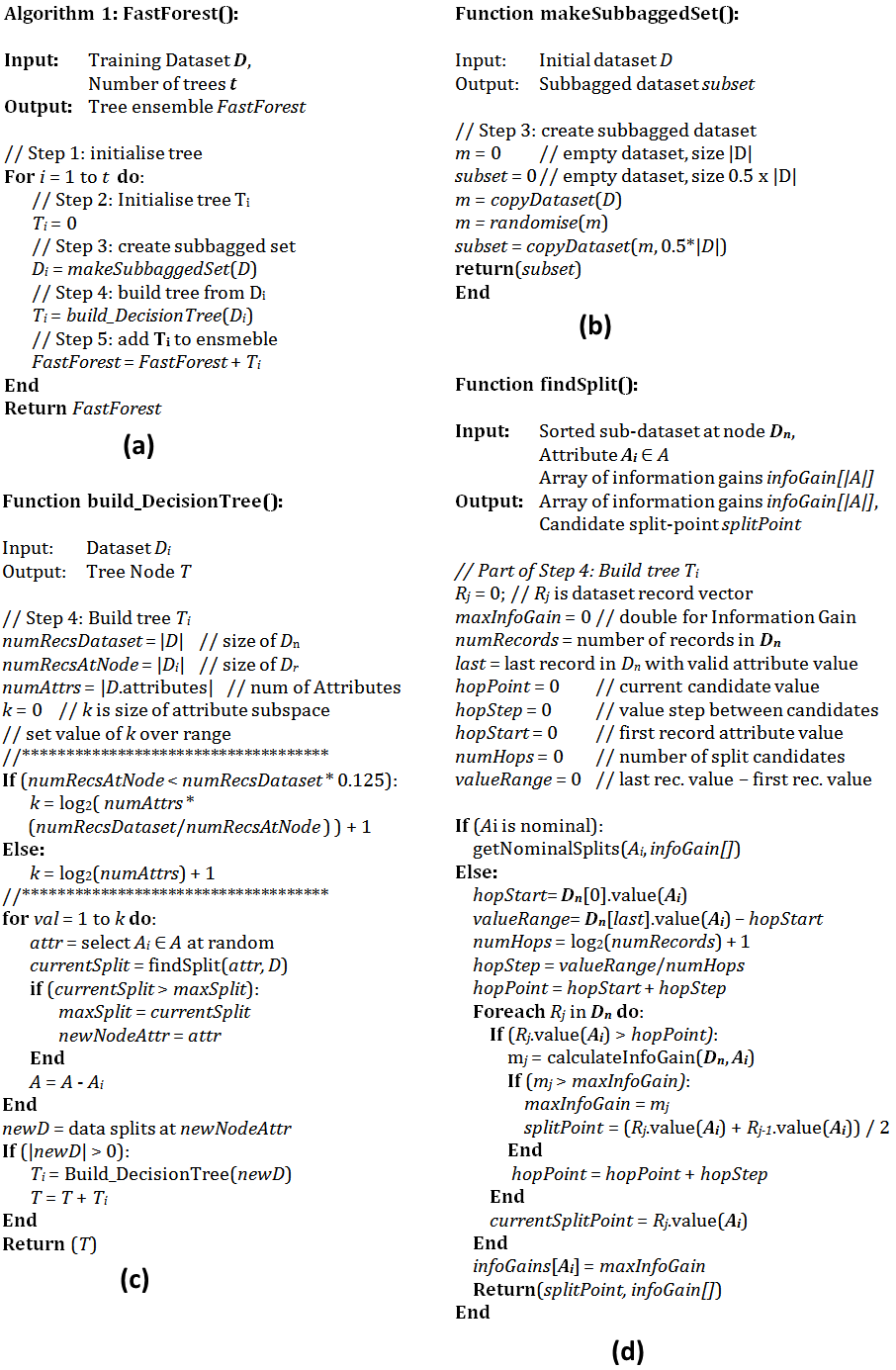}
\caption{Algorithm code for (a) the FastForest algorithm, (b) the subbagging component, (c) the tree-building function, incorporating Dynamic Restricted Subspacing (DRS) and (d) Logarithmic Split-Point Sampling (LSPS) components.}
\end{figure}

\subsection{Proposed FastForest Algorithm}
The FastForest algorithm, presented in Fig. 7, executes the following steps:\\
\begin{itemize}
\item \textbf{Step 1: initialise the tree loop (Fig. 7a)} - this creates a for-loop based on the user-selected number of trees required, $t$. For each run of the loop, the following steps are executed:\\
\item \textbf{Step 2: initialise tree $T_i$  (Fig. 7a)} - this sets up a tree object $T_i$ into which, the $i$-th tree of $|T|$ trees will be stored.\\
\item \textbf{Step 3: create a subbagged dataset (Fig. 7b)} - this calls the `makeSubbaggedSet()' function to make a subbagged dataset. This is done by making a copy of the initial training dataset $D$, randomising it, then returning the first half of the records as the subbagged dataset.\\
\item \textbf{Step 4: build a tree from the subbagged dataset (Fig. 7c and 7d)} - the subbagged dataset created in Step 3 is passed to the `Build\_DecisionTree()' function and a tree is created. The process executes recursively, with the function calling itself as it builds the tree node-by-node. In the process, the attribute subsample $k$ is set by the Dynamic Restricted Subspacing (DRS) component. Once $k$ is set, all $k$ attributes are tested as possible new node candidates. At this time, the `Build\_DecisionTree()' function calls the `findSplit()' function (Fig. 7d). This implements the Logarithmic Split-Point Sampling (LSPS) component to efficiently select the split-point candidates and identifies the attribute split-point that returns the maximum information gain of the values tested. Once built, the tree is returned and stored in tree object $T_i$.\\
\item \textbf{Step 5: add the tree to the forest (Fig. 7a)} - the tree $T_i$ just completed in Step 4 is now added to the FastForest ensemble object. At this point, the for-loop created in Step 1 now loops back to Step 2 to create a new subbagged dataset and tree. This loop completes $|T|$ runs, at which time, the FastForest object is returned and cast as output from the FastForest() algorithm. It can now be used to classify similar previously-unseen records.  
\end{itemize}

\section{Experiments}

To quantify the effect each component has on classification accuracy and processing speed, Section 4 tests each component, both individually and combined, as well as in comparison with other popular ensemble classifiers, before a final comparison with Random Forest itself. However, this section begins with an identification of the datasets used for testing.

\begin{table}
\centering
\caption{Details of datasets in Group 1, featuring 15 numerical attributes (all non-class attributes are numerical).}
\label{tab:1}       
\begin{tabular}{lllllll}
\hline\noalign{\smallskip}
ID & Dataset & Abbrev. & Records & Attrs & Classes & Type \\
\noalign{\smallskip}\hline\noalign{\smallskip}
1&	Banknote&BN&	1,372&	5&	2 &Num.	\\
2&	Ecoli&EC&	336&	8&	2&Num.\\
3&	Ionosphere&ION&	351&	35&	2&Num.\\
4&	Libra M'ments&LIB&	360&	91&	15&Num.\\
5&	Liver Disorder&LIV&	345&	7&	2&Num.\\
6&	Mfeat-fourier&MFOU&	2,000&	77&	10&Num.\\
7&	Mfeat-zernike&MZER&	2,001&	48&	10&Num.\\
8&	OBS Network&OBS&	1,075&	22&	4&Num.\\					
9&	QSAR Biodeg.&QSAR&	1,051&	41&	2&Num.		\\
10&	Seeds&SEE&	210&	8&	3&Num.\\
11&	Segment&SEG&	2,310&	20&	7&Num.\\
12&	Seismic-bumps&SEIS&	2,584&	19&	2&Num.\\
13&	Thyroid&THY&	215&	6&	3&Num.\\
14&	Urban Land Cover&ULC&	675&	148&	9&Num.\\
15&	Yeast&YEA&	1,484&	9&	10&Num.\\
\noalign{\smallskip}\hline
\end{tabular}
\end{table}

\begin{table}
\centering
\caption{Details of datasets in Group 2, featuring 15 categorical (Cat.) and mixed (Mix.) attributes.}
\label{tab:1}       
\begin{tabular}{lllllll}
\hline\noalign{\smallskip}
ID & Dataset & Abbrev. & Records & Attrs & Classes & Type  \\
\noalign{\smallskip}\hline\noalign{\smallskip}
16&	Abalone&	ABA &4,177&	8&	29&	Mix.\\
17&	Annealing	&ANN&798	&39&	6&	Mix\\
18&	Arrh’mia&	ARR& 452	&280&	16&	Mix.\\
19&	Balance Scale&BAL&	625&	5&	3&	Cat.\\
20&	Car Evaluation&CAR&	1,728&	7&	4&	Cat.\\
21&	Chronic Kidney Disease&CKD	&400&	25&	2&	Mix.\\
22&	CMC	&CMC&1,473	&10&	3&	Mix.\\
23&	Hepatitis&	HEP&155&	20&	2&	Mix.\\					
24&	Hypothyroid&HYP	&3,772	&30&	4&	Mix.\\
25&	KRKPA7&KRK	&3,196&	37&	2&	Cat.\\
26&	Soybean&SOY&	683	&36&	19&	Cat.\\
27&	SPECT&SPE&	267	&23	&2&	Cat.\\
28&	Splice&SPL&	3,190&	62&	3&	Cat.\\
29&	TicTacToe&TIC&	958&	10&	2&	Cat.\\
30&	Web Phishing&WEB&	1,353&	10&	3&	Cat.\\

\noalign{\smallskip}\hline
\end{tabular}
\end{table}

\begin{table}
\centering
\caption{Details of datasets in Group 3, featuring 15 large datasets.}
\label{tab:1}       
\begin{tabular}{lllllll}
\hline\noalign{\smallskip}
ID & Dataset & Abbrev. & Records & Attrs & Classes & Type  \\
\noalign{\smallskip}\hline\noalign{\smallskip}
31&	Adult	&ADU&48,842&	14&	41&	Mix.	\\
32&	Bank&BAN&	45,211	&17	&2&	Mix.	\\
33&	Credit Card&CC&	30,000&	24&	2&	Num.	\\
34&	C-Map&CMAP&	10,545&	29&	6&	Num.	\\
35&	El. Board&	EB&45,781&	5&	31	&Mix.	\\
36&	HTRU-2&HTR&	17,898&	9&	2&	Num.\\
37&	MAGIC Telescope&MT&	19,020&	11&	2&	Num.	\\
38&	MNIST&MNI&	10,000&	785&	10&	Num.	\\
39&	MOCAP&MOC&	78,095&	38&	5&	Num.\\
40&	Mushroom	&MUS&8,124&	23&	2&	Cat.\\
41&	Nursery&NUR&	12,960&	9&	5&	Cat.\\
42&	Pen-digits&PEN&	10,992&	17&	10	&Num.\\
43&	Sensorless Drive&SSD&	58,509&	49	&11&	Num.\\
44&	Shuttle&SHU&	43,500&	9&	7&	Num.\\
45&	Skin Segmentation&SS&	245,057&	4	&2&	Num.\\
\noalign{\smallskip}\hline
\end{tabular}
\end{table}

\indent To validate the proposed methods in Section 3, a total of 45 datasets in three groups were utilised during testing. Group 1 consists of 15 datasets with non-class numerical attributes, Group 2 features 15 datasets with mixed (numerical/categorical) non-class attributes and Group 3 includes a mix of 15 larger datasets. These datasets, listed in Tables 2 to 4, respectively, are from the University of California Irvine (UCI) data repository \cite{RN464}.

\subsection{Processing Speed of FastForest on PC Hardware}
The combined effects of the three proposed components of FastForest on processing speed were tested by recording model build times on all 30 datasets of Groups 1 and 2. These were then compared with five ensemble classifiers - SysFor \cite{RN466}, Random Subspace \cite{RN456}, ForestPA \cite{RN468}, Bagging \cite{RN455} and Random Committee \cite{RN522}. All testing was performed on an Intel quad-core Core i5 PC with Windows operating system, 16GB of RAM, Intel 535-series solid-state drive and Weka version 3.8.3 (this version implements multi-core support and was enabled for testing). All tests were timed on three occasions, with an average of the three recorded.\\
\indent FastForest achieved the fastest cumulative build time over the 30 datasets, completing models in a combined 2.78 seconds, well ahead of ForestPA at 73.21 seconds and twice as fast as the 5.6 seconds of Random Subspace. In the process, FastForest achieved 24 speed wins out of the 30 datasets, followed by SysFor with five and Random Subspace with two.

\begin{table}
\centering
\caption{Processing speed times (lower is better) of FastForest against five other ensemble classifiers from the Weka implementation using the 15 numerical datasets of Group 1}
\label{tab:1}       
\begin{tabular}{llp{1.1cm}p{1.1cm}p{1.1cm}p{1.1cm}p{1.1cm}p{1.1cm}}
\hline\noalign{\smallskip}
&&Bagging & Random Committee& Random Subspace & SysFor & ForestPA & FastForest  \\
ID. & Dataset & Time (secs) & Time (secs) & Time (secs) & Time (secs) & Time (secs) & Time (secs)\\
\noalign{\smallskip}\hline\noalign{\smallskip}
&Wins & 0 & 0 & 1 & 1 & 0 &\textbf{13}  \\
\noalign{\smallskip}\hline\noalign{\smallskip}
1&	BN	&0.045&	0.075&	0.048&	0.172&	0.204&	\textbf{0.028}\\
2&	EC	&0.018&	0.024&	0.022&	0.016&	0.038&	\textbf{0.014}\\
3&	ION&	0.078&	0.057&	0.056&	0.395&	0.200&	\textbf{0.028}\\
4&	LIB&	0.537&	0.117&	0.382&	1.544&	1.512&	\textbf{0.053}\\
5&	LIV&	0.031&	0.032&	0.027&	0.141&	0.072&	\textbf{0.019}\\
6&	MFOU&	2.656&	0.748&	1.724&	12.565&	8.182&	\textbf{0.293}\\
7&	MZER&	1.792&	0.688&	1.152&	8.125&	5.746&	\textbf{0.269}\\
8&	OBS&	0.176&	0.108&	0.107&	0.399&	0.477&	\textbf{0.050}\\
9&	QSAR&	0.343&	0.144&	0.208&	1.648&	0.692&	\textbf{0.074}\\
10&	SEE&	0.022&	0.025&	0.025&	0.052&	0.037&	\textbf{0.008}\\
11&	SEG&	0.380&	0.259&	0.271&	1.581&	1.288&	\textbf{0.119}\\
12&	SEIS&	0.380&	0.272&	0.209&	\textbf{0.118}	&1.184&	0.145\\
13&	THY&	0.012&	0.022&	0.019&	0.032&	0.027&	\textbf{0.010}\\
14&	ULC&	1.114&	0.201&	0.690&	4.912&	2.982&	\textbf{0.081}\\
15&	YEA	&0.198&	0.232&	\textbf{0.134}&	0.388&	0.917&	0.143\\
\noalign{\smallskip}\hline\noalign{\smallskip}
&Total (secs)&	7.781&	3.003&	5.073&	32.088&	23.557&	\textbf{1.331}\\
\noalign{\smallskip}\hline
\end{tabular}
\end{table}

\begin{table}
\centering
\caption{Processing speed times (lower is better) of FastForest against five other ensemble classifiers using the 15 categorical and mixed datasets of Group 2.}
\label{tab:1}       
\begin{tabular}{llp{1.1cm}p{1.1cm}p{1.1cm}p{1.1cm}p{1.1cm}p{1.1cm}}
\hline\noalign{\smallskip}
&&Bagging & Random Committee & Random Subspace  & SysFor & ForestPA & FastForest  \\
ID. & Dataset & Time (secs) & Time (secs) & Time (secs) & Time (secs) & Time (secs) & Time (secs)\\
\noalign{\smallskip}\hline\noalign{\smallskip}
&Wins &0&	0&	1&	4	&0	&\textbf{11}  \\
\noalign{\smallskip}\hline\noalign{\smallskip}
16&	ABA	&1.152&	1.319&	0.756&	4.251	&10.524	&\textbf{0.713}\\
17&	ANN	&0.079	&0.034&	0.064&	\textbf{0.014}&	0.563&	0.019\\
18&	ARR	&1.423&	0.177	&0.862&	0.176	&3.590&	\textbf{0.089}\\
19&	BAL&	0.028&	0.037&	\textbf{0.022}&	0.072&	0.113&	\textbf{0.022}\\
20&	CAR&	0.049&	0.027&	0.036&	0.028&	0.873&	\textbf{0.021}\\
21&	CKD&	0.044&	0.044&	0.033&	\textbf{0.016}&	0.233&	0.017\\
22&	CMC	&0.137	&0.117&	0.090&	0.274&	1.335&	\textbf{0.074}\\
23&	HEP&	0.026&	0.043&	0.076&	0.031&	0.081&	\textbf{0.010}\\
24&	HYP&	0.380&	0.260&	0.474	&\textbf{0.009}&	3.441&	0.171\\
25&	KRK&	0.535&	0.181&	0.441	&0.559&	3.164&	\textbf{0.083}\\
26&	SOY&	0.183&	0.083&	0.114&	\textbf{0.006}&	2.513&	0.053\\
27&	SPE&	0.032&	0.027&	0.035&	0.040&	0.148&	\textbf{0.015}\\
28&	SPL&	0.987&	0.181&	0.630&	0.872&	21.864&	\textbf{0.117}\\
29&	TIC&	0.058&	0.031&	0.043&	0.046&	0.436&	\textbf{0.018}\\
30&	WEB&	0.069&	0.040&	0.050&	0.035&	0.779&	\textbf{0.024}\\
\noalign{\smallskip}\hline\noalign{\smallskip}
&Total (secs)&	5.181&	2.601&	3.726&	6.429	&49.656	&\textbf{1.447}\\
\noalign{\smallskip}\hline
\end{tabular}
\end{table}

\begin{table}
\centering
\caption{Classification accuracy (higher is better) of FastForest against five other ensemble classifiers using the 15 numerical datasets of Group 1 (highest scores bolded).}
\label{tab:7}       
\begin{tabular}{llp{1.15cm}p{1.15cm}p{1.15cm}p{1.15cm}p{1.15cm}p{1.15cm}}
\hline\noalign{\smallskip}
&&Bagging & Random Committee & Random Subspace  & SysFor  & ForestPA & FastForest  \\
ID. & Dataset &Acc'y (\%) & Acc'y (\%) & Acc'y (\%) & Acc'y (\%) & Acc'y (\%) & Acc'y (\%)\\
\noalign{\smallskip}\hline\noalign{\smallskip}
&Wins &1&	4&	3&	1&	1&	\textbf{8}\\
\noalign{\smallskip}\hline\noalign{\smallskip}
1&	BN&	98.91&	98.98&	97.30&	98.47&	99.42&	\textbf{99.56}\\
2	&EC&	90.18&	88.10&	88.99&	\textbf{91.67}&	88.99&	89.58\\
3&	ION&	92.02&	93.16&	\textbf{94.02}&	91.74&	91.74&	93.45\\
4&	LIB&	78.33&	\textbf{83.89}&	77.22&	65.56&	72.78&	81.94\\
5&	LIV&	73.33&	69.57&	69.86&	66.96&	69.86&	\textbf{75.94}\\
6&	MFOU&	81.05&	84.30&	83.40&	79.40&	80.00&	\textbf{84.5}\\
7&	MZER&	74.35&	\textbf{78.05}&	76.75&	73.85&	75.30&	77.95\\
8&	OBS&	99.63&	\textbf{100.00}&	\textbf{100.00}&	92.74&	99.63&	\textbf{100.00}\\
9&	QSAR&	86.54&	\textbf{87.39}&	85.40&	83.41&	85.02&	86.73\\
10&	SEE	&91.90&	92.86&	90.95&	88.10&	89.52&	\textbf{95.24}\\
11&	SEG&	96.84&	98.18&	96.75&	95.54&	96.58&	\textbf{98.66}\\
12&	SEIS&	93.38&	92.84&	\textbf{93.42}&	93.38&	\textbf{93.42}&	93.11\\
13&	THY&	93.95	&93.95&	93.49&	93.95&	91.63&	\textbf{95.81}\\
14&	ULC&	85.19&	86.67&	84.74&	81.93&	82.67&	\textbf{86.81}\\
15&	YEA	&\textbf{62.13}	&60.24&	61.05&	58.22&	60.65&	61.32\\
\noalign{\smallskip}\hline\noalign{\smallskip}
&Avg. (\%)&	86.52&	87.21	&86.22&	83.66&	85.15&	\textbf{88.04}\\
\noalign{\smallskip}\hline
\end{tabular}
\end{table}

\begin{table}
\centering
\caption{Classification accuracy (higher is better) of FastForest against five other ensemble classifiers using the 15 categorical and mixed datasets from Group 2.}
\label{tab:1}       
\begin{tabular}{llp{1.15cm}p{1.15cm}p{1.15cm}p{1.15cm}p{1.15cm}p{1.15cm}}
\hline\noalign{\smallskip}
&&Bagging & Random Committee & Random Subspace  & SysFor  & ForestPA  & FastForest  \\
ID. & Dataset &Acc'y (\%) & Acc'y (\%) & Acc'y (\%) & Acc'y (\%) & Acc'y (\%) & Acc'y (\%)\\
\noalign{\smallskip}\hline\noalign{\smallskip}
&Wins &3&	2&	3&	0&	3&	\textbf{6}\\
\noalign{\smallskip}\hline\noalign{\smallskip}
16&	ABA&	25.33&	23.25&	\textbf{26.74}&	23.99&	24.63&	23.96\\
17&	ANN&	97.41&	\textbf{99.35}&	98.22&	94.98&	98.70&	\textbf{99.35}\\
18&	ARR&	\textbf{74.34}&	69.03&	71.46&	51.55&	73.89&	66.81\\
19&	BAL&	\textbf{84.64}&	77.76&	82.88&	80.00&	83.36&	82.24\\
20&	CAR&	92.42&	94.68&	70.02&	88.25&	\textbf{97.16}&	94.21\\
21&	CKD&	99.25&	99.75&	\textbf{100.00}&	94.25&	99.75&	\textbf{100.00}\\
22&	CMC&	54.31&	51.32&	\textbf{54.85}&	54.58&	54.18&	52.07\\
23&	HEP&	82.58&	83.87&	81.94&	81.94&	85.16&	\textbf{87.10}\\
24&	HYP&	99.55&	99.60&	94.88&	92.29&	\textbf{99.63}&	98.63\\
25&	KRK&	99.06&	\textbf{99.41}&	97.34&	98.03&	99.09&	99.28\\
26&	SOY&	86.24&	93.41&	90.78&	74.08&	93.41&	\textbf{93.56}\\
27&	SPE&	\textbf{83.15}&	80.15&	79.40&	82.77&	82.77&	79.40\\
28&	SPL&	94.20&	95.96&	96.11&	91.16&	95.58&	\textbf{96.68}\\
29&	TIC&	93.63&	96.66&	82.57&	86.33&	\textbf{97.39}&	97.18\\
30&	WEB&	89.50&	89.36&	85.29&	88.47&	89.73&	\textbf{89.95}\\

\noalign{\smallskip}\hline\noalign{\smallskip}
&Avg. (\%)&	83.71&	83.57&	80.83&	78.84&	\textbf{84.96}&	84.04\\
\noalign{\smallskip}\hline
\end{tabular}
\end{table}
However, when the datasets are split according to their initial groupings, FastForest is more dominant with numerical datasets, winning 13 of 15 Group-1 tests, as shown in Table 5 (winning scores bolded). On a cumulative build time basis with numerical datasets, FastForest delivered the fastest result at 1.331 seconds, followed by Random Committee at just on three seconds. \\
\indent Testing with the 15 mixed datasets of Group 2, as shown in Table 6, revealed FastForest to be less dominant. However, it still managed 11 wins out of 15 datasets, whilst achieving the fastest cumulative build time at 1.44 seconds, with Random Committee the next-fastest at 2.60 seconds. Out of the 11 wins for FastForest, one was a tie with Random Subspace, while SysFor achieved four wins. Thus, at this level, FastForest achieves superior processing speed results compared with a number of popular and recent ensemble classifiers.

\subsection{Classification Accuracy of FastForest on PC Hardware}
While classification accuracy was not the primary goal of FastForest, any significant loss in accuracy would make using it a more difficult choice. Table 7 shows the comparison results between FastForest and the other ensemble classifiers, testing for classification accuracy on the 15 numerical datasets of Group 1. All tests were implemented with ten-fold cross-validation. \\
\indent In this setting, FastForest outperformed the other five classifiers, winning eight of 15 tests, with one tie. The nearest competitor was RandomCommittee with four wins, while Random Subspace won three. In achieving the most wins, FastForest also delivered the highest average accuracy across the Group 1 datasets, scoring 88.04\%, followed by Random Committee, on 87.21\%.\\
\indent However, classification accuracy on the mixed datasets of Group 2 proved more complex, as Table 8 shows. Again, FastForest achieved the most individual wins, with six, followed by Bagging, Random Subspace and ForestPA all scoring three wins each. Although ForestPA achieved only half the number of wins of FastForest, ForestPA's overall average classification result was slightly higher at 84.9\%, with FastForest placing second with 84.0\%. However, the issue facing ForestPA is that the time required to achieve these results on this `mixed' group of datasets is more than 33 times greater than for FastForest (ForestPA required 49.65 seconds against FastForest's 1.48 seconds).\\
\indent Moreover, this slightly lower categorical result was not unexpected - by design, the Logarithmic Split-Point Sampling (LSPS) component of FastForest cannot work with categorical attributes, thus FastForest only uses subbagging and DRS on these attribute types. Still, FastForest achieved twice as many wins as ForestPA and a much higher overall classification average than Random Subspace. Combine this with a much faster processing speed than either of these alternatives and FastForest continues to compare well.

\subsection{Component testing}
While the results comparing FastForest with other ensemble classifiers are broadly promising, it is also important to quantify how each of the three components contributes to classification accuracy and processing speed individually. This section details testing implemented to identify each component's contribution to accuracy and speed gains across the 30 datasets from groups 1 and 2.

\begin{figure}
\centering
\includegraphics[scale=0.3]{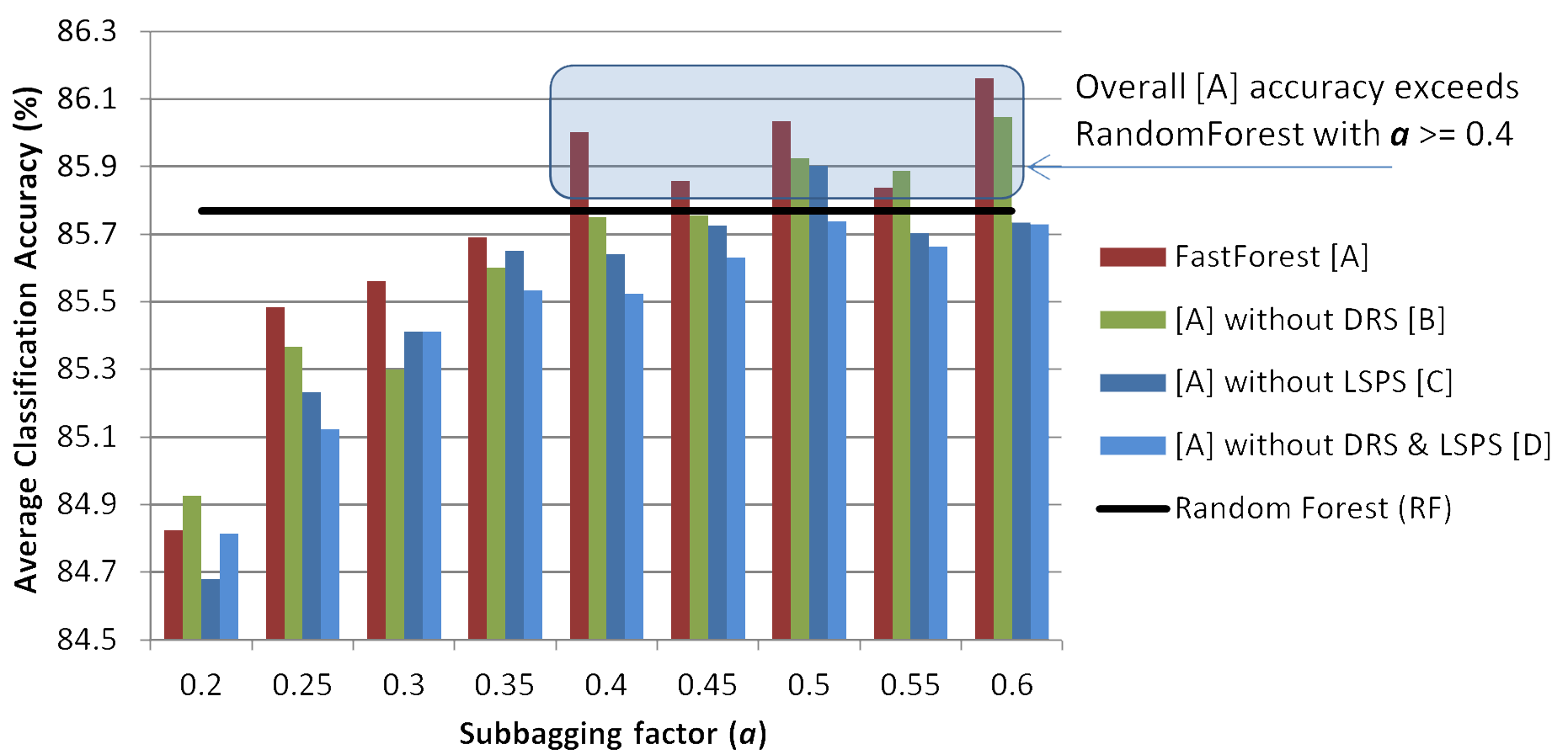}
\caption{Component testing for accuracy (higher is better) of FastForest on the Group 1 and 2 datasets.}
\end{figure}
\begin{figure}
\centering
\includegraphics[scale=0.3]{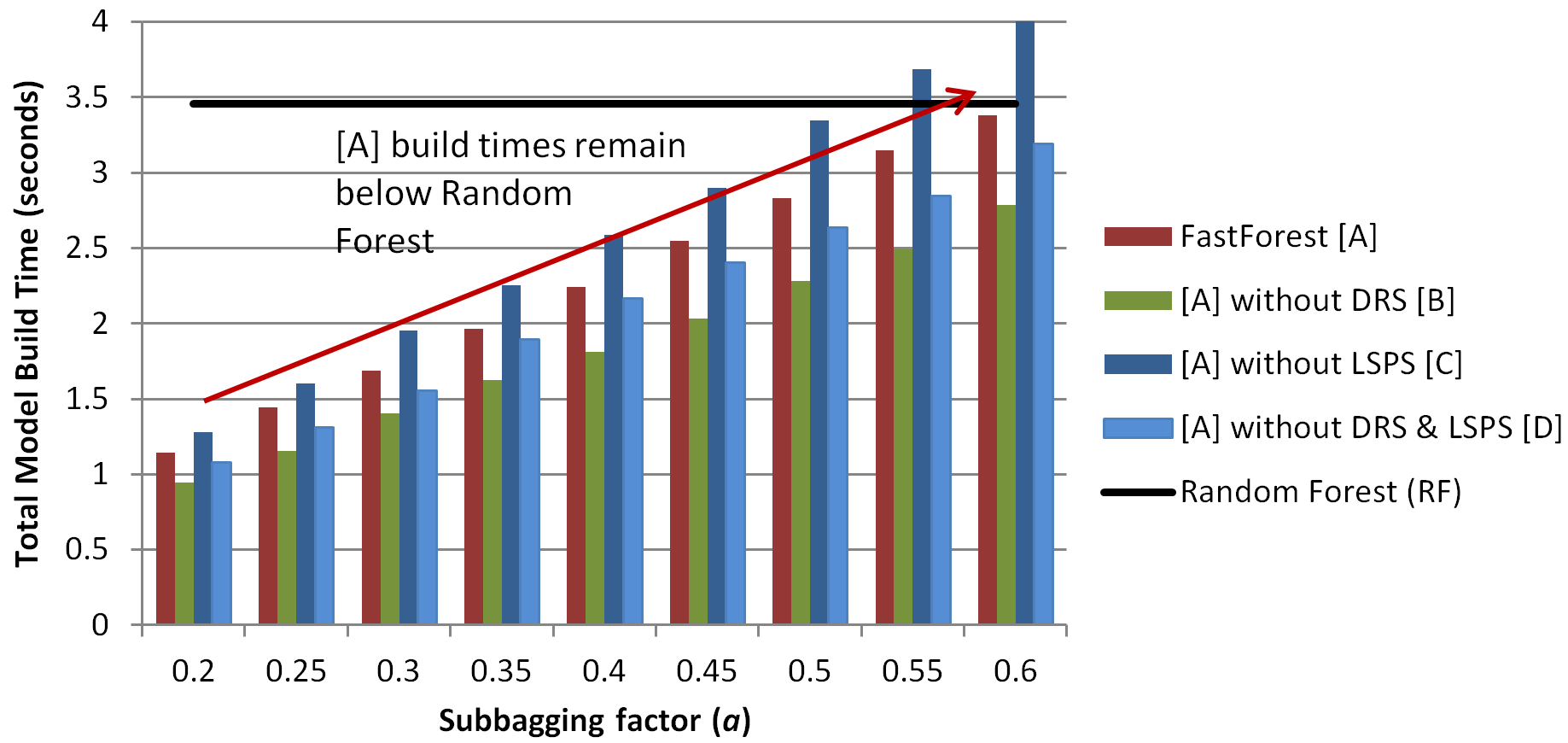}
\caption{Component testing for speed (lower is better) of FastForest on the Group 1 and 2 datasets.}
\end{figure}
The results in Figs. 8 and 9 show the effect of components being removed from the FastForest algorithm and are compared directly against Random Forest across the 30 test datasets of Groups 1 and 2 for average classification accuracy and total build time, respectively. The horizontal axis of each graph is the subbagging scale factor, ranging from 0.2 to 0.6 in 0.05-increments. Thus, the results also show accuracy and build time as a function of subbagging factor $a$, as discussed in Section 3.1. The results of Fig. 8 compare the average classification accuracy of the complete FastForest algorithm (Results [A]) against versions of FastForest with various components removed. For example, results [B] show average accuracy for FastForest without DRS (Section 3.3) and results [C], average accuracy for FastForest without LSPS (Section 3.2). Results [D] remove both DRS and LSPS, leaving FastForest with the subbagging component only. The results [A], [B], [C] and [D] in Fig. 9 show the effect of the same component removal process on total build times. Thus, the difference between [A] and [B] is the effect of DRS, between [A] and [C] the effect of LSPS.
\subsubsection{Subbagging}
From a broad perspective, changes in the subbagging scale factor have a clear effect on both average classification accuracy (see Fig. 8) and model build times (Fig. 9). Further, they also broadly match the patterns in Fig. 4. While overall accuracy exceeds Random Forest in Fig. 8 with a minimum subbagging scale factor of just 0.4, a factor of 0.5 provides the highest accuracy result for subbagging alone and just below that of Random Forest (Fig. 8 [D]). Moreover, this $a$=0.5 factor remains well below Random Forest for processing speed in Fig. 9. Thus, an $a$=0.5 subbagging factor is an effective base for FastForest.
\subsubsection{Logarithmic Split-Point Sampling (LSPS)}
The consistent rise in build times between results [A] and [C] in Fig. 9 shows the clear negative effect of removing LSPS, with its inclusion improving build times by a significant margin. This drop in build times due to the use of LSPS also occurs regardless of the subbagging scale factor. Moreover, this is not unexpected - the reduction in the number of split-point candidates tested as a result of LSPS also reduces the overall processing workload. Moreover, Fig. 8 shows that the `[B] vs [D]' results indicate LSPS also provides an improvement in classification accuracy across eight of nine subbagging scale factors without the presence of DRS. 
\subsubsection{Dynamic Restricted Subspacing (DRS)}
The importance of DRS to FastForest is shown by the differences in the [A] and [B] results in both Figs. 8 and 9. The benefit of DRS on average accuracy is clear from Fig. 8. The weakest accuracy result at each subbagging scale factor occurs with subbagging alone [D]. The inclusion of LSPS alone not only improves build times across the board in Fig. 9 ([B] vs [D]), but also improves average classification accuracy in eight of the nine subbagging groups in Fig.8. However, DRS improves upon LSPS further, achieving a higher average classification accuracy result again in seven of nine subbagging scale factors in Fig. 8. Moreover, the improvement is to such an extent that if the subbagging factor is at least $a$=0.4, FastForest delivers greater average classification accuracy than Random Forest across the Group 1 and 2 datasets.\\
\indent Thus, while DRS incurs a small cost in processing time ([A] vs [B], Fig. 9) regardless of the subbagging factor used, some of that expense is offset by the gains achieved through LSPS ([A] vs [C]). Without LSPS, the overall build times [A] would be proportionally greater. Importantly, these overall FastForest build times with DRS on-board are still below the time achieved by Random Forest with standard bagging (RF). 
As a result, the time penalty incurred by DRS is more than compensated for by the gains in classification accuracy resulting from DRS. This enables FastForest to, on average, exceed Random Forest, yet do so in less time. For the record, FastForest achieved its greatest accuracy result with a scalar of 0.6, where it achieved a small accuracy gain of 0.39\% over Random Forest for the Group 1 and 2 dataset tests, but in less time than Random Forest (see Figs. 8 and 9).
\subsubsection{Quantifying the effect of DRS over Dynamic Subspacing}
We have seen in Figs. 8 and 9 how Dynamic Restricted Subspacing (DRS) positively affects classification accuracy albeit with a slight cost in processing speed, but how does DRS compare with the previous `dynamic subspacing' method? To identify the effect, all three dataset groups, totalling 45 datasets, were used. The Group 3 datasets have higher record counts, ranging from 8,124 to 245,057. Each group was tested on FastForest in three configurations - no dynamic subspacing at all [A], the full `dynamic subspacing' (DS) algorithm \cite{RN451} [B] and our proposed DRS method [C]. These results are summarised in Table 9.

\begin{table}
\centering
\caption{Comparison of DRS options on the all three groups of datasets, for average accuracy (higher is better) and total build time (lower is better).}
\label{tab:1}       
\begin{tabular}{p{1.5cm}p{0.85cm}p{0.85cm}p{0.85cm}p{0.85cm}p{0.85cm}p{0.85cm}p{0.85cm}p{0.85cm}p{0.85cm}}
\hline\noalign{\smallskip}
&Random Forest&&FastForest&&FastForest && FastForest  \\
&(Reference)&&No DRS/DS [A]&&DS [B]&& DRS [C]\\
Dataset Group& Acc. (\%)&Time (s)&Acc. (\%)&Time (s)&Acc. (\%)&Time (s)&Acc. (\%)&Time (s)\\
\noalign{\smallskip}\hline\noalign{\smallskip}
Group 1&	87.66&	1.75&	88.01&	1.14&	87.87&	1.43&	\textbf{88.04}&	\textbf{1.13}\\
Group 2&	83.89&	1.68&	83.83&	\textbf{1.16}&	83.88&	1.54&	\textbf{84.03}&	1.44\\
Group 3&	\textbf{93.44}&	125.13&	93.36&	\textbf{76.17}&	93.41&	100.61&	93.42&	94.12\\
\noalign{\smallskip}\hline\noalign{\smallskip}
Avg/Total	&88.33A&	128.56T&	88.40A	&\textbf{78.47T}&	88.39A&	103.58T&	\textbf{88.50A}&	96.69T\\
\noalign{\smallskip}\hline
\end{tabular}
\end{table}

FastForest in all three subspacing options in Table 9 achieved higher average classification accuracy and faster build times over all 45 datasets than Random Forest. Overall, the fastest build times were achieved with no additional subspacing included (A), but, again, this is not unexpected, since any form of dynamic subspacing increases the number of attributes and, thus, the processing workload. However, the proposed DRS component (C) achieved the highest average classification accuracy of the three subspacing options across all test groups and in less time than the previous dynamic subspacing method.\\
Moreover, as Table 9 also shows and will now be discussed in detail in Section 4.4, FastForest achieved a time saving advantage of just over 31 seconds, a 24\% speed increase over Random Forest. It also did this with an overall average classification accuracy gain of 0.17\% over RandomForest. 

\begin{figure}
\centering
\includegraphics[scale=0.35]{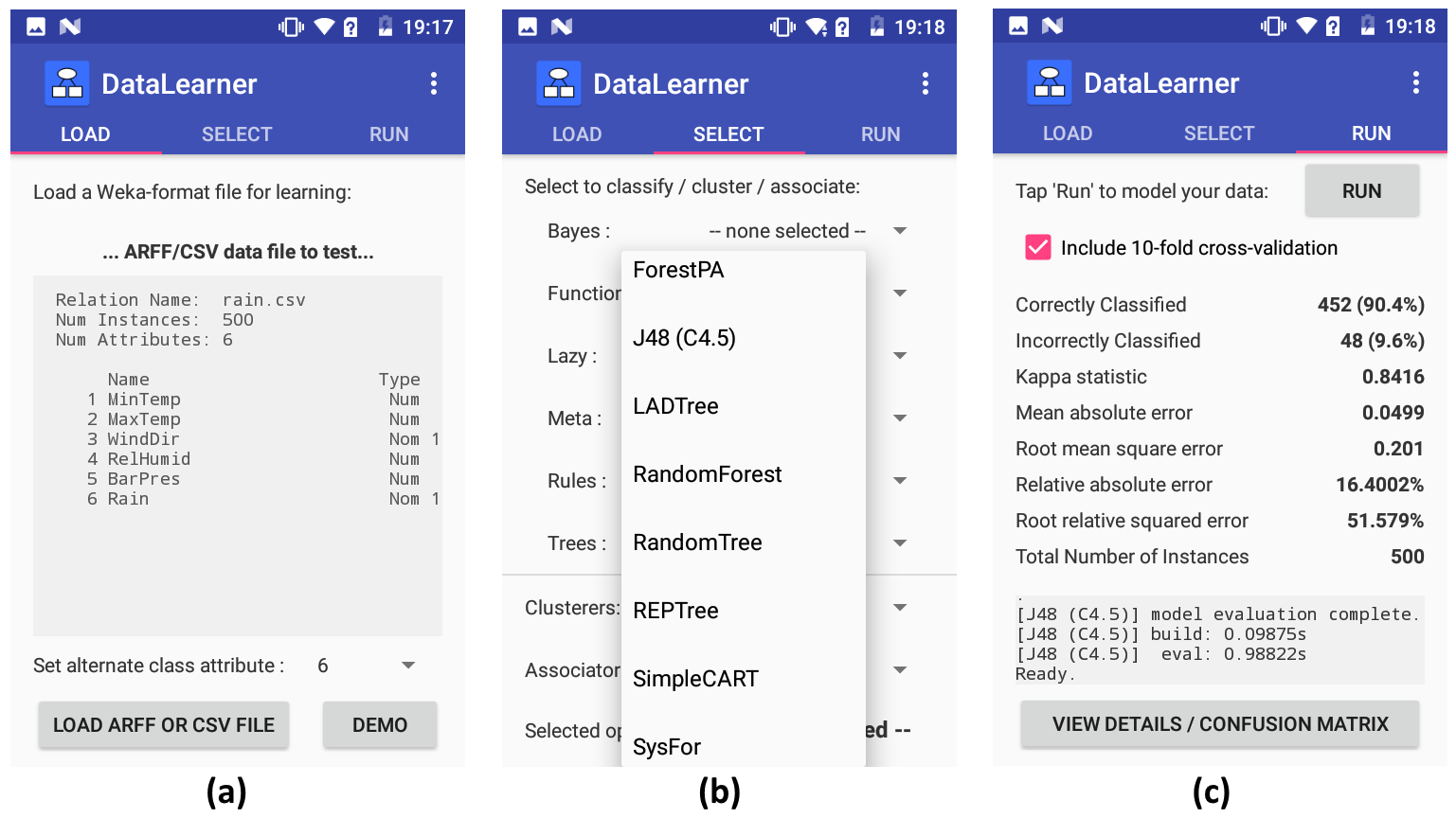}
\caption{FastForest has also been developed for a future update of the DataLearner app \cite{RN598} for Android devices (current version shown operating on a budget Plus8-brand Touch4 Android smartphone).}
\end{figure}

\subsection{Comparison with Random Forest}
The last question to be answered in this paper is how FastForest compares with Random Forest. Reasons for this research discussed in Section 1 are the arrival of smartphones, the Internet of Things and the need for efficient algorithms, along with the continued popularity of ensemble classifiers. To this end, two versions of FastForest were developed. The first was for Weka version 3.8.3 on PC (the test PC featured Intel Core i5 CPU, 16GB of RAM, Windows 8.1 operating system and Intel 535-series solid state drive). The second was for an un-released update of DataLearner \cite{RN598} for smartphone use (as shown in Fig. 10, tested on a 2019 Samsung Galaxy A30 phone with 8-core CPU, 3GB of RAM and Android 9.0 operating system). The current version of DataLearner is freely available on Google Play.

\begin{table}
\centering
\caption{Accuracy (higher is better) and speed (lower is better) of FastForest vs RandomForest for Group 1 datasets on PC and smartphone.}
\label{tab:1}       
\begin{tabular}{p{1.65cm}p{0.85cm}p{0.85cm}p{0.85cm}p{0.85cm}p{0.85cm}p{0.85cm}p{0.85cm}p{0.85cm}p{0.85cm}}
\hline\noalign{\smallskip}
& \multicolumn{4}{l}{Windows 8.1 (Weka 3.8.3)}&\multicolumn{4}{l}{Galaxy A30 (DataLearner 1.1.7)} \\
& \multicolumn{2}{l}{Accuracy (\%) }& \multicolumn{2}{l}{Speed (secs)} &\multicolumn{2}{l}{Accuracy (\%) }& \multicolumn{2}{l}{Speed (secs)} \\
\hline\noalign{\smallskip}
Dataset & Rand. Forest & Fast Forest & Rand. Forest & Fast Forest  & Rand. Forest & Fast Forest & Rand. Forest & Fast Forest \\
\hline\noalign{\smallskip}
Wins/Ties &4/4&	\textbf{7/4}&	2/2&	\textbf{11/2}&	3/2&\textbf{10/2}&0/0&\textbf{15/0}\\
\noalign{\smallskip}\hline\noalign{\smallskip}
1 BN&	99.20&	\textbf{99.56}&	0.042&	\textbf{0.028}&		99.34&	\textbf{99.71}&	1.268&	\textbf{0.608}\\
2 EC	&88.99&	\textbf{89.58}&	\textbf{0.013}&	0.014&		89.29&	\textbf{90.18}&	0.312&	\textbf{0.228}\\
3 ION&	92.88&	\textbf{93.45}&	0.028&	\textbf{0.021}	&	93.16&	\textbf{93.73}&	0.823&	\textbf{0.470}\\
4 LIB&	81.94&	81.94&	0.073&	\textbf{0.053}&		\textbf{83.89}&	82.22	&1.936&	\textbf{1.215}\\
5 LIV&	73.04&	\textbf{75.94}&	0.019&	0.019&		\textbf{73.33}&	73.04&	0.544&	\textbf{0.415}\\
6 MFOU&	84.50&	84.50&	0.444&	\textbf{0.293}	&	83.20&	\textbf{83.65}&	12.791&	\textbf{6.770}\\
7 MZER&	\textbf{78.00}&	77.95&	0.391&	\textbf{0.269}&		76.70&	\textbf{77.70}	&11.281&	\textbf{6.442}\\
8 OBS&	100.00&	100.00&	0.057&	\textbf{0.050}&		100.00&	100.00&	1.876&	\textbf{1.283}\\
9 QSAR&	\textbf{87.01}&	86.73&	0.082&	\textbf{0.074}&		85.97&	\textbf{87.58}&	2.898&	\textbf{1.980}\\
10 SEE&	94.29&	\textbf{95.24}&	0.016&	\textbf{0.008}&		93.81&	\textbf{94.29}&	0.246&	\textbf{0.165}\\
11 SEG&	97.88&	\textbf{98.66}&	0.161&	\textbf{0.119}&		97.92&	\textbf{98.53}&	4.942&	\textbf{2.807}\\
12 SEIS&	\textbf{93.38}&	93.11&	0.148&	\textbf{0.145}&		93.11&	\textbf{93.27}&	5.883&	\textbf{4.154}\\
13 THY&	94.88&	\textbf{95.81}&	\textbf{0.013}&	0.014&		95.35&	\textbf{96.74}&	0.200&	\textbf{0.132}\\
14 ULC&	86.81&	86.81&	0.121&	\textbf{0.081}&		\textbf{87.11}&	86.81&	3.341&	\textbf{1.863}\\
15 YEA	&\textbf{62.13}&	61.32&	0.143&	0.143&		61.12&	61.12&	4.713&	\textbf{3.892}\\
\noalign{\smallskip}\hline\noalign{\smallskip}
AVG/Total&	87.66&	\textbf{88.04}&	1.75&	\textbf{1.33}	&	87.55&	\textbf{87.91}&	53.05&	\textbf{32.42}\\
\noalign{\smallskip}\hline
\end{tabular}
\end{table}

\begin{table}
\centering
\caption{Accuracy and speed of FastForest vs RandomForest (PC and smartphone) on Group 2 datasets (accuracy: higher is better, speed: lower is better, best scores bolded) }
\label{tab:1}       
\begin{tabular}{p{1.55cm}p{0.85cm}p{0.85cm}p{0.85cm}p{0.85cm}p{0.85cm}p{0.85cm}p{0.85cm}p{0.85cm}p{0.85cm}}
\hline\noalign{\smallskip}
& \multicolumn{4}{l}{Windows 8.1 (Weka 3.8.3)}&\multicolumn{4}{l}{Galaxy A30 (DataLearner 1.1.7)} \\
& \multicolumn{2}{l}{Accuracy (\%) }& \multicolumn{2}{l}{Speed (secs)} &\multicolumn{2}{l}{Accuracy (\%) }& \multicolumn{2}{l}{Speed (secs)} \\
\hline\noalign{\smallskip}
Dataset & Rand. Forest & Fast Forest & Rand. Forest & Fast Forest  & Rand. Forest & Fast Forest & Rand. Forest & Fast Forest \\
\noalign{\smallskip}\hline\noalign{\smallskip}
Wins/ Ties      &          	5/3&	\textbf{7/3}&	2/1&	\textbf{12/1}	&	6/3&	6/3	&0/0	&\textbf{15/0}\\
\noalign{\smallskip}\hline\noalign{\smallskip}
16 ABA&\textbf{24.13}&	23.96	&0.798&	\textbf{0.713}	&	24.44&	\textbf{24.54}&	22.980&	\textbf{18.614}\\
17 ANN &	99.35&	99.35&	0.019&	0.019&		99.35&	99.35&	0.546&	\textbf{0.388}\\
18 ARR	&\textbf{67.26}&	66.81&	0.109&	\textbf{0.089}&		\textbf{69.03}&	67.04&	2.648&	\textbf{1.971}\\
19 BSC&	81.44&	\textbf{82.24}&	0.023&	\textbf{0.022}	&	81.92&	\textbf{82.40}&	0.782&	\textbf{0.507}\\
20 CE&	\textbf{94.50}&	94.21&	0.023&	\textbf{0.021}	&	\textbf{94.50}&	93.81&	0.698&	\textbf{0.480}\\
21 CKD	&100.00	&100.00&	0.039&	\textbf{0.017}&		99.75&	99.75&	0.495&	\textbf{0.319}\\
22 CMC&	51.46&	\textbf{52.07}&	\textbf{0.071}&	0.074	&	\textbf{51.66}&	51.46&	2.233&	\textbf{1.910}\\
23 HEP	&85.16&	\textbf{87.10}&	0.042&	\textbf{0.010}	&	84.52&	84.52&	0.394&	\textbf{0.314}\\
24 HYP	&\textbf{99.31}&	98.62&	\textbf{0.158}&	0.171	&	\textbf{99.36}&	98.67&	5.189&	\textbf{3.871}\\
25 KRK&	99.22&	\textbf{99.28}	&0.119&	\textbf{0.083}&		99.12&	\textbf{99.41}&	3.736&	\textbf{2.193}\\
26 SOY	&92.97&	\textbf{93.56}&	0.075&	\textbf{0.053}&		92.09&	\textbf{93.12}&	1.808&	\textbf{1.233}\\
27 SPE&	\textbf{81.27}&	79.40&	0.020&	\textbf{0.015}	&	\textbf{81.27}&	79.78&	0.286&	\textbf{0.206}\\
28 SPS&	95.92&	\textbf{96.68}&	0.127&	\textbf{0.117}	&	95.67&	\textbf{96.46}&	3.095&	\textbf{2.589}\\
29 TIC	&96.35&	\textbf{97.18}&	0.025&	\textbf{0.018}&		96.76&	\textbf{97.18}&	0.596&	\textbf{0.437}\\
30 WEB&	89.95&	89.95&	0.032	&\textbf{0.024}&		\textbf{97.34}&	97.20&	12.347&	\textbf{8.199}\\
\noalign{\smallskip}\hline\noalign{\smallskip}
AVG/Total	&83.89&	\textbf{84.03}&	1.68	&\textbf{1.45}	&	\textbf{84.45}&	84.31&	57.83	&\textbf{43.23}\\
\noalign{\smallskip}\hline
\end{tabular}
\end{table}

\begin{table}
\centering
\caption{PC results of FastForest against Random Forest against the 15 larger datasets of Group 3.}
\label{tab:13}       
\begin{tabular}{lp{0.75cm}p{0.65cm}p{0.7cm}p{0.7cm}p{1cm}p{1cm}p{1cm}p{1cm}p{1cm}}
\hline\noalign{\smallskip}
&&&&& \multicolumn{2}{l}{Accuracy (\%) }&\multicolumn{2}{l}{Speed (secs)}\\
Dataset &Records& Attrs & Classes & Type & Random Forest & Fast Forest & Random Forest & Fast Forest\\
\noalign{\smallskip}\hline\noalign{\smallskip}
Wins  &&&&&7&	7&	1&	\textbf{14}\\
\noalign{\smallskip}\hline\noalign{\smallskip}
31 ADU&	48842&	14	&41&	Mix&	\textbf{91.735}&	91.485	&4.687	&\textbf{4.432}\\
32 BAN	&45211&	17&	2&	Mix&	\textbf{90.396}&	90.199&	3.934&	\textbf{3.634}\\
33 CC	&30000&	24	&2&	Num&	81.663&	\textbf{81.770}&	\textbf{7.160}&	7.652\\
34 CMAP&	10545&	29&	6&	Num&	\textbf{94.908}&	94.642&	1.685&	\textbf{1.218}\\
35 EB	&45781&	5	&31&	Mix&	65.584&	\textbf{65.730}&	7.963&	\textbf{3.622}\\
36 HTRU	&17898	&9&	2	&Num	&97.994	&\textbf{98.011}&	1.944&	\textbf{1.548}\\
37 MT	&19020&	11&	2&	Num&	88.002&	\textbf{88.076}&	3.193&	\textbf{2.532}\\
38 MNI&	10000	&785&	10&	Num&	\textbf{95.330}&	95.160&	4.119	&\textbf{3.526}\\
39 MOC&	78095&	38&	5&	Num	&97.927&	\textbf{98.225}&	51.188&	\textbf{34.966}\\
40 MUSH&	8124	&23	&2&	Cat&	100.000&	100.000&	0.127&	\textbf{0.083}\\
41 NURS&	12960&	9	&5	&Cat&	\textbf{99.066}&	98.943&	0.213&	\textbf{0.159}\\
42 PEN	&10992	&17	&10&	Num	&99.127	&\textbf{99.190}&	1.152&	\textbf{0.951}\\
43 SSD	&58509&	49&	11	&Num&	\textbf{99.899}&	99.865&	19.749&	\textbf{13.960}\\
44 SHU	&43500&	9	&7&	Num	&\textbf{99.984}&	99.982&	4.181&	\textbf{3.151}\\
45 SS	&245057	&4	&2	&Num&	99.958	&\textbf{99.961}&	13.839&	\textbf{12.687}\\
\noalign{\smallskip}\hline\noalign{\smallskip} 
Avg./Total	&& 	 	&&& 	 	\textbf{93.44}A&	93.42A	&125.13T	&\textbf{94.12}T\\
\noalign{\smallskip}\hline
\end{tabular}
\end{table}

The smartphone tests were conducted on the 30 datasets of Groups 1 and 2, while the PC tests were carried out on all three test groups introduced in Section 4. DataLearner \cite{RN598} is currently limited to single-core processing and thus, testing with the large record-count datasets of Group 3 would have been excessive. However, DataLearner \cite{RN598} is the only application of its type on Google Play at time of writing and in its first iteration. \\
\indent Table 10 highlights the results of the 15 numerical datasets of Group 1 and shows a dominant outcome on both PC and Android platforms. The Windows result saw FastForest achieve 11 speed wins to two against Random Forest, with two ties. Moreover, FastForest also delivered seven wins for classification accuracy to Random Forest's four, as well as achieving a small overall average accuracy gain of 0.38\%. The smartphone results followed suit, with FastForest comprehensively winning all on all 15 datasets for processing speed, while also taking the accuracy honours against Random Forest 10 datasets to three.\\ 
\indent Table 11 shows the results of the 15 mixed datasets of Group 2 on both platforms. The Windows results show FastForest achieving six times as many speed wins as Random Forest, scoring 12 wins to two. In the process, FastForest also outscored Random Forest on accuracy, seven wins to five. The smartphone tests again followed a similar trend, with FastForest clearly taking the lead for processing speed, winning all 15 tests. At the same time, FastForest showed it could still match Random Forest for classification accuracy, with each algorithm claiming six wins and three ties.\\
\indent Over the two groups of smartphone tests in Tables 10 and 11, FastForest achieved a clear-cut advantage over Random Forest for processing speed, recording a comprehensive 30-0 win-loss score. Concurrently, FastForest also delivered a 16-9 win-loss score over Random Forest for classification accuracy, with five ties over the 30 datasets of Groups 1 and 2. FastForest also achieved a higher overall accuracy average of 86.11\% to Random Forest's 86.00\%. \\
\indent The results in Table 12 of the 15 larger datasets in Group 3 tested on the Windows platform reveal FastForest again maintaining a considerable advantage over Random Forest for processing speed, achieving 14 wins to Random Forest's one and an overall speed gain of 24\%. As with the results of Group 2 in Table 11, both FastForest and Random Forest shared a 7-1 win-tie record for classification accuracy. Overall, Random Forest recorded a narrow lead for overall average classification accuracy, 93.44\% to FastForest's 93.42\%. \\
\indent Across the three dataset groups, the Windows results reveal FastForest achieved twice as many accuracy `wins' as Random Forest, scoring 28 to 14, with three ties. FastForest also achieved a slightly higher overall average classification accuracy of 88.49\% versus 88.33\%. However, while maintaining (and even exceeding) classification accuracy is important, the main goal of FastForest's development was model build speed. In this regard, FastForest clearly out-paced Random Forest, scoring 37 wins to five, with an overall build time of 96.9 seconds to Random Forest's 128.5 seconds, a speed gain of 24\%.
\subsection{Reasons for platform result differences}
With the same techniques being applied to both the Windows and smartphone versions of FastForest, it is reasonable to assume they would also produce the same classification accuracy results, if not the same processing speed. However, the differences are not the result of CPU architectural difference between PCs and smartphones, but rather different implementations of the Weka data-mining suite. First, Weka 3.8.3 used on the Windows tests has support for multi-core processors and both Random Forest and FastForest take advantage of this. In contrast, the current version of the DataLearner app \cite{RN598} uses Weka version 3.6.15 by necessity and only supports single-core processing. Thus, to ensure comparability with the available version of Weka on each platform (and their Random Forest implementations), separate versions of FastForest were built using each version's Random Forest implementation as a base.\\
\indent Moreover, further analysis of the source code for the Random Forest implementation within the two Weka releases indicates additional differences in the way the Bagging function operates. In both cases, Random Forest acts as a `meta' classifier, that is, it directs one or more secondary classifiers to provide the basic functions. Inside Weka 3.6.15, the Bagging process follows the traditional method whereby each dataset resample $D_j$ consists of $n$ records, including record duplicates. However, the Bagging code inside Weka 3.8.3 appears to utilise a weighting system where $n$ samples are taken from the dataset $D$, but only those samples that are unique become part of the bagged dataset $D_j$, $|D_j|$ $\neq$ n. This has the effect of reducing the size of each dataset bootstrap to approximately 63.2\% of $|D|$. However, it also reduces the processing time during the tree induction process for each $D_j$. 
\section{Conclusion}
Through its combination of Bagging and Random Subspace techniques, Random Forest continues to be amongst the top ensemble classifiers in terms of classification accuracy and processing speed. Moreover, it also continues to feature consistently in new research. However, with the ubiquitousness of smartphones and the rise of the Internet of Things, in conjunction with new data mining platforms such as TensorFlow Lite and DataLearner, data mining on constrained hardware is seeing growing demand. Thus, the need to continue improving the processing speed of ensemble classifiers, such as Random Forest, whilst maintaining their high accuracy is paramount.\\
\indent This paper proposes FastForest, an optimisation of Random Forest that delivers an overall tested speed gain of over 24\% compared with RandomForest on testing using 45 datasets. At the same time, FastForest matched and frequently exceeded Random Forest for classification accuracy. These tests further showed that FastForest won more than seven times as many dataset processing speed tests as it lost to Random Forest, while also winning twice as many tests for classification accuracy. Moreover, FastForest proved to be faster than a range of other ensemble classifiers, including more than 33 times faster than ForestPA and as much as four-times faster than Bagging. \\
\indent This research discussed in detail the three components - subbagging, Logarithmic Split-Point Sampling (LSPS) and Dynamic Restricted Subspacing (DRS) - that deliver these results. Moreover, tests carried out on an Android smartphone using 30 datasets proved the efficacy of FastForest in mobile environments, out-pacing Random Forest on all 30 datasets, whilst also matching Random Forest in classification accuracy.  \\
\indent With multi-core processors in seemingly limitless supply and available in everything from PCs to microcontrollers, techniques requiring the sequential induction of trees would appear to be at a disadvantage in terms of speed. This leaves room for further research into improving identification of an optimum tree count during the forest induction process.


\ifCLASSOPTIONcompsoc
  \section*{Acknowledgments}
\else
  \section*{Acknowledgment}
\fi
This research is supported by an Australian Government Research Training Program (RTP) scholarship.

\bibliographystyle{IEEEtran}

\begin{thebibliography}{10}
\providecommand{\url}[1]{#1}
\csname url@samestyle\endcsname
\providecommand{\newblock}{\relax}
\providecommand{\bibinfo}[2]{#2}
\providecommand{\BIBentrySTDinterwordspacing}{\spaceskip=0pt\relax}
\providecommand{\BIBentryALTinterwordstretchfactor}{4}
\providecommand{\BIBentryALTinterwordspacing}{\spaceskip=\fontdimen2\font plus
\BIBentryALTinterwordstretchfactor\fontdimen3\font minus
  \fontdimen4\font\relax}
\providecommand{\BIBforeignlanguage}[2]{{%
\expandafter\ifx\csname l@#1\endcsname\relax
\typeout{** WARNING: IEEEtran.bst: No hyphenation pattern has been}%
\typeout{** loaded for the language `#1'. Using the pattern for}%
\typeout{** the default language instead.}%
\else
\language=\csname l@#1\endcsname
\fi
#2}}
\providecommand{\BIBdecl}{\relax}
\BIBdecl

\bibitem{RN519}
H.-J. Zhu, T.-H. Jiang, B.~Ma, Z.-H. You, W.-L. Shi, and L.~Cheng, ``Hemd: a
  highly efficient random forest-based malware detection framework for
  android,'' \emph{Neural Computing and Applications}, vol.~30, no.~11, pp.
  3353--3361, 2018.

\bibitem{RN520}
E.~V. Sylvester, P.~Bentzen, I.~R. Bradbury, M.~Clément, J.~Pearce, J.~Horne,
  and R.~G. Beiko, ``Applications of random forest feature selection for
  fine‐scale genetic population assignment,'' \emph{Evolutionary
  applications}, vol.~11, no.~2, pp. 153--165, 2018.

\bibitem{RN529}
Z.~Wang, Y.~Wang, R.~Zeng, R.~S. Srinivasan, and S.~Ahrentzen, ``Random forest
  based hourly building energy prediction,'' \emph{Energy and Buildings}, vol.
  171, pp. 11--25, 2018.

\bibitem{RN455}
L.~Breiman, ``Bagging predictors,'' \emph{Machine learning}, vol.~24, no.~2,
  pp. 123--140, 1996.

\bibitem{RN456}
I.~Barandiaran, ``The random subspace method for constructing decision
  forests,'' \emph{IEEE Trans. Pattern Anal. Mach. Intell}, vol.~20, no.~8, pp.
  1--22, 1998.

\bibitem{RN454}
L.~Breiman, ``Random forests,'' \emph{Machine learning}, vol.~45, no.~1, pp.
  5--32, 2001.

\bibitem{RN544}
E.~M. Silveira, S.~H.~G. Silva, F.~W. Acerbi-Junior, M.~C. Carvalho, L.~M.~T.
  Carvalho, J.~R.~S. Scolforo, and M.~A. Wulder, ``Object-based random forest
  modelling of aboveground forest biomass outperforms a pixel-based approach in
  a heterogeneous and mountain tropical environment,'' \emph{International
  Journal of Applied Earth Observation and Geoinformation}, vol.~78, pp.
  175--188, 2019.

\bibitem{RN545}
P.~Probst, M.~N. Wright, and A.~Boulesteix, ``Hyperparameters and tuning
  strategies for random forest,'' \emph{Wiley Interdisciplinary Reviews: Data
  Mining and Knowledge Discovery}, vol.~9, no.~3, p. e1301, 2019.

\bibitem{RN546}
H.~Ishwaran and M.~Lu, ``Standard errors and confidence intervals for variable
  importance in random forest regression, classification, and survival,''
  \emph{Statistics in medicine}, vol.~38, no.~4, pp. 558--582, 2019.

\bibitem{RN547}
L.~Benali, G.~Notton, A.~Fouilloy, C.~Voyant, and R.~Dizene, ``Solar radiation
  forecasting using artificial neural network and random forest methods:
  Application to normal beam, horizontal diffuse and global components,''
  \emph{Renewable energy}, vol. 132, pp. 871--884, 2019.

\bibitem{RN548}
S.~Lakshmanaprabu, K.~Shankar, M.~Ilayaraja, A.~W. Nasir, V.~Vijayakumar, and
  N.~Chilamkurti, ``Random forest for big data classification in the internet
  of things using optimal features,'' \emph{International Journal of Machine
  Learning and Cybernetics}, pp. 1--10, 2019.

\bibitem{RN533}
\BIBentryALTinterwordspacing
n.d., ``Class randomforest,'' n.d. [Online]. Available:
  \url{http://weka.sourceforge.net/doc.dev/weka/classifiers/trees/RandomForest.html}
\BIBentrySTDinterwordspacing

\bibitem{RN598}
D.~Yates, M.~Z. Islam, and J.~Gao, ``Datalearner: a data mining and knowledge
  discovery tool for android smartphones and tablets,'' in \emph{International
  Conference on Advanced Data Mining and Applications}.\hskip 1em plus 0.5em
  minus 0.4em\relax Springer, 2019, Conference Proceedings, pp. 828--838.

\bibitem{RN496}
\BIBentryALTinterwordspacing
J.~Saarinen, ``Aws to switch to per-second billing for linux instances,'' 2017.
  [Online]. Available:
  \url{https://www.itnews.com.au/news/aws-to-switch-to-per-second-billing-for-linux-instances-473600}
\BIBentrySTDinterwordspacing

\bibitem{RN461}
D.~Yates, M.~Z. Islam, and J.~Gao, ``Spaarc: A fast decision tree algorithm,''
  in \emph{Australasian Conference on Data Mining}.\hskip 1em plus 0.5em minus
  0.4em\relax Springer, 2019, Conference Proceedings, pp. 43--55.

\bibitem{RN451}
M.~N. Adnan and M.~Z. Islam, ``Effects of dynamic subspacing in random
  forest,'' in \emph{International Conference on Advanced Data Mining and
  Applications}.\hskip 1em plus 0.5em minus 0.4em\relax Springer, 2017,
  Conference Proceedings, pp. 303--312.

\bibitem{RN538}
L.~Rokach, ``Ensemble-based classifiers,'' \emph{Artificial Intelligence
  Review}, vol.~33, no. 1-2, pp. 1--39, 2010.

\bibitem{RN537}
K.~Tumer and J.~Ghosh, ``Error correlation and error reduction in ensemble
  classifiers,'' \emph{Connection science}, vol.~8, no. 3-4, pp. 385--404,
  1996.

\bibitem{RN468}
M.~N. Adnan and M.~Z. Islam, ``Forest pa: Constructing a decision forest by
  penalizing attributes used in previous trees,'' \emph{Expert Systems with
  Applications}, vol.~89, pp. 389--403, 2017.

\bibitem{RN497}
J.~Jia, Z.~Liu, X.~Xiao, B.~Liu, and K.-C. Chou, ``ippi-esml: an ensemble
  classifier for identifying the interactions of proteins by incorporating
  their physicochemical properties and wavelet transforms into pseaac,''
  \emph{Journal of theoretical biology}, vol. 377, pp. 47--56, 2015.

\bibitem{RN536}
J.~J. Rodriguez, L.~I. Kuncheva, and C.~J. Alonso, ``Rotation forest: A new
  classifier ensemble method,'' \emph{IEEE transactions on pattern analysis and
  machine intelligence}, vol.~28, no.~10, pp. 1619--1630, 2006.

\bibitem{RN539}
J.~Han, J.~Pei, and M.~Kamber, \emph{Data mining: concepts and
  techniques}.\hskip 1em plus 0.5em minus 0.4em\relax Elsevier, 2011.

\bibitem{RN462}
L.~Breiman, \emph{Classification and regression trees}.\hskip 1em plus 0.5em
  minus 0.4em\relax Routledge, 2017.

\bibitem{RN463}
J.~R. Quinlan, \emph{C4. 5: programs for machine learning}.\hskip 1em plus
  0.5em minus 0.4em\relax Elsevier, 2014.

\bibitem{RN549}
------, ``Induction of decision trees,'' \emph{Machine learning}, vol.~1,
  no.~1, pp. 81--106, 1986.

\bibitem{RN498}
E.~Bauer and R.~Kohavi, ``An empirical comparison of voting classification
  algorithms: Bagging, boosting, and variants,'' \emph{Machine learning},
  vol.~36, no. 1-2, pp. 105--139, 1999.

\bibitem{RN470}
P.~Buhlmann and B.~Yu, ``Analyzing bagging,'' \emph{The Annals of Statistics},
  vol.~30, no.~4, pp. 927--961, 2002.

\bibitem{RN524}
J.~H. Friedman and P.~Hall, ``On bagging and nonlinear estimation,''
  \emph{Journal of statistical planning and inference}, vol. 137, no.~3, pp.
  669--683, 2007.

\bibitem{RN499}
F.~Zaman and H.~Hirose, ``Effect of subsampling rate on subbagging and related
  ensembles of stable classifiers,'' in \emph{International Conference on
  Pattern Recognition and Machine Intelligence}.\hskip 1em plus 0.5em minus
  0.4em\relax Springer, 2009, Conference Proceedings, pp. 44--49.

\bibitem{RN525}
P.~Buhlmann, \emph{Bagging, boosting and ensemble methods}.\hskip 1em plus
  0.5em minus 0.4em\relax Springer, 2012, pp. 985--1022.

\bibitem{RN599}
\BIBentryALTinterwordspacing
F.~Eibe, M.~A. Hall, and I.~H. Witten, \emph{The WEKA Workbench. Online
  Appendix for "Data Mining, Practical Machine Learning Tools and
  Techniques"}.\hskip 1em plus 0.5em minus 0.4em\relax Morgan Kaufmann, 2016,
  vol.~4. [Online]. Available:
  \url{https://www.cs.waikato.ac.nz/ml/weka/Witten_et_al_2016_appendix.pdf}
\BIBentrySTDinterwordspacing

\bibitem{RN452}
P.~Latinne, O.~Debeir, and C.~Decaestecker, ``Limiting the number of trees in
  random forests,'' in \emph{International workshop on multiple classifier
  systems}.\hskip 1em plus 0.5em minus 0.4em\relax Springer, 2001, Conference
  Proceedings, pp. 178--187.

\bibitem{RN453}
T.~M. Oshiro, P.~S. Perez, and J.~A. Baranauskas, ``How many trees in a random
  forest?'' in \emph{International workshop on machine learning and data mining
  in pattern recognition}.\hskip 1em plus 0.5em minus 0.4em\relax Springer,
  2012, Conference Proceedings, pp. 154--168.

\bibitem{RN535}
H.~Grahn, N.~Lavesson, M.~H. Lapajne, and D.~Slat, ``Cudarf: a cuda-based
  implementation of random forests,'' in \emph{2011 9th IEEE/ACS International
  Conference on Computer Systems and Applications (AICCSA)}.\hskip 1em plus
  0.5em minus 0.4em\relax IEEE, 2011, Conference Proceedings, pp. 95--101.

\bibitem{RN530}
S.~Bernard, L.~Heutte, and S.~Adam, ``Influence of hyperparameters on random
  forest accuracy,'' in \emph{International Workshop on Multiple Classifier
  Systems}.\hskip 1em plus 0.5em minus 0.4em\relax Springer, 2009, Conference
  Proceedings, pp. 171--180.

\bibitem{RN464}
\BIBentryALTinterwordspacing
D.~Dua and C.~Graff, ``Uci machine learning repository,'' 2019. [Online].
  Available: \url{http://archive.ics.uci.edu/ml}
\BIBentrySTDinterwordspacing

\bibitem{RN466}
Z.~Islam and H.~Giggins, ``Knowledge discovery through sysfor: a systematically
  developed forest of multiple decision trees,'' in \emph{Proceedings of the
  Ninth Australasian Data Mining Conference-Volume 121}.\hskip 1em plus 0.5em
  minus 0.4em\relax Australian Computer Society, Inc., 2011, Conference
  Proceedings, pp. 195--204.

\bibitem{RN522}
\BIBentryALTinterwordspacing
n.d., ``Class randomcommittee,'' n.d. [Online]. Available:
  \url{http://weka.sourceforge.net/doc.dev/weka/classifiers/meta/RandomCommittee.html}
\BIBentrySTDinterwordspacing

\end{thebibliography}


\end{document}